\documentclass[11pt]{article}

\usepackage[final]{acl}

\usepackage{times}
\usepackage{latexsym}

\usepackage[T1]{fontenc}

\usepackage[utf8]{inputenc}

\usepackage{microtype}

\usepackage{inconsolata}

\usepackage{graphicx}

\usepackage{booktabs}
\usepackage{multirow}

\title{More Choices, Fewer Decisions: Ordinal-Scale Bias in JEV-like Direct-Decision Models}

\author{Tianxiang Gao$^{1,2}$ \quad
        Jinzhe Li$^{1,4}$ \quad 
        Zhiyuan Li$^{1}$ \quad
        Yi Chang$^{1,3,4}$ \quad   
        Yuan Wu$^{1}$\thanks{Corresponding author} \\
        $^{1}$School of Artificial Intelligence, Jilin University \\
        $^{2}$ College of Software, Jilin University \\
        $^{3}$Engineering Research Center of Knowledge-Driven Human-Machine Intelligence, Jilin University \\
        $^{4}$International Center of Future Science, Jilin University\\
         \{gaotx5524, jinzhe25, zhiyuanl24\}@mails.jlu.edu.cn, \{yichang, yuanwu\}@jlu.edu.cn \\
  }

\begin{document}
\maketitle
\begin{abstract}
Direct-decision models enable efficient classification and evaluation by selecting from predefined candidates, but their reliability depends on more than prediction accuracy.
Across 40 datasets and nine models from the JEV, KEV, and StartLux Decision families, we identify a systematic tendency to underuse ordinal rating scales.
We call this phenomenon \emph{ordinal scale-utilization bias}: final decisions concentrate on fewer effective labels than the gold distribution.
The bias is substantially weaker on tasks with unordered semantic categories.
Evaluating the same items on finer ordinal scales strengthens compression despite controls for gold imbalance and candidate position.
Randomizing candidate order reduces but does not eliminate compression, while widening ordinal gaps at fixed candidate count can improve accuracy without restoring utilization.
For most models, concentrated decisions coexist with broad average candidate probabilities.
Greater model size improves accuracy without consistently restoring scale utilization.
Together, these findings characterize compression as strongly associated with ordinal structure and intensified by scale refinement, rather than fully explained by presentation position or local spacing.
This bias can be effectively mitigated through targeted training, making scale utilization an actionable reliability target alongside accuracy.
The code and data are available at
\url{https://anonymous.4open.science/r/jev_ordinal_scale_bia/}.
\end{abstract}

\section{Introduction}

Language models increasingly convert text into fixed labels, grades, and rubric scores \citep{zhang2015character,borkan2019nuanced,liu-etal-2023-g,zheng2023judging}.
Direct-decision models such as JEV and the open KEV family return a distribution over supplied candidates without generating a textual verdict \citep{typesafe2026api,palmer2026kev}.
Their structured outputs and low latency make them attractive classifiers and evaluators \citep{rao2026jevrubric,li2026jevjudge}.
Reliable use, however, requires both correct item-level decisions and faithful use of the supplied scale.

\begin{figure}[t]
  \centering
  \includegraphics[width=\columnwidth]{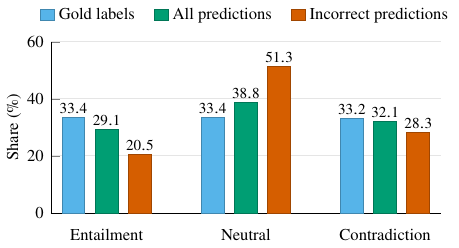}
  \caption{\label{fig:anli}
    \textbf{JEV~1.13 label shares on ANLI.} Gold-label and all-prediction shares use all 6,400 development and test items; incorrect-prediction shares use the 1,603 errors.
  }
\end{figure}

Candidate sets may be \emph{nominal}, with distinct unordered categories, or \emph{ordinal}, with labels arranged along a graded scale.
Ordinal levels encode neighborhood and distance information that nominal multiclass treatment discards \citep{frank-hall-2001-ordinal,gutierrez-etal-2016-ordinal}.
A model can attain plausible accuracy while repeatedly selecting only a small part of that scale: accuracy records item-level correctness, not population-level scale use.
Prior work documents label, order, and judge-specific preferences \citep{zhao2021calibrate,fei-etal-2023-mitigating,zheng2024mcq,zheng2023judging,wang-etal-2024-large-language-models-fair}, but systematic underuse of ordinal scales remains unresolved.

Our starting point is ANLI \citep{nie-etal-2020-adversarial}.
JEV~1.13 reaches 74.95\% accuracy on its 6,400 development and test items, yet assigns 38.8\% of all predictions and 51.3\% of its 1,603 errors to Neutral (Figure~\ref{fig:anli}).
This skew is not copied from the data: the gold labels are nearly balanced at 2,140 Entailment, 2,136 Neutral, and 2,124 Contradiction.
Nor is it tied to a fixed slot: Neutral occupies each candidate position for about one third of the items and is selected for 37.2--39.6\% at each position.
Because ANLI is nominal, Neutral may be a special attractor; the pattern is therefore a clue rather than evidence of ordinal compression.
It nevertheless exposes the central problem: plausible accuracy can coexist with a systematically distorted output distribution.

We call the broader behavior \emph{ordinal scale-utilization bias}: final decisions systematically occupy fewer effective ordinal levels than the gold labels in the same population.
Its measurable signature, \emph{candidate-space compression}, compares their entropy-based effective supports \citep{hill1973diversity,jost2006entropy}.
This separates scale use from nearby explanations: low accuracy concerns correctness, gold imbalance describes the evaluation population, position bias reflects presentation sensitivity, and candidate count describes granularity.
None alone establishes faithful use of an ordinal scale.

We test this account with JEV~1.13, three KEV models, and five StartLux Decision models.
Across 36 ordinal datasets, decisions use only 60--76\% of effective gold support, versus 87--102\% on four nominal tasks (\S\ref{sec:broad}).
Candidate-order randomization weakens but does not remove the gap (\S\ref{sec:position}).
On fixed items with balanced gold support and positions, utilization falls as $K$ grows from 2 to 14 even though probabilities remain broad (\S\ref{sec:kscaling}).
At fixed candidate count, widening ordinal gaps can improve accuracy without restoring candidate utilization (\S\ref{sec:distance}).
This dissociation shows that local ordinal proximity alone does not explain compression.
Finally, targeted post-training recovers scale utilization on supervised KEV tasks and reveals model-size-dependent transfer to unseen sources (\S\ref{sec:posttraining}).

The main contributions of this paper are threefold:
\begin{itemize}
  \item \textbf{Diagnosing:} We formulate ordinal scale-utilization bias as a population-level reliability problem, measure absolute effective-support deviation from the gold labels, and document compression across 36 ordinal datasets and nine models, contrasted with four nominal tasks.
  \item \textbf{Disentangling:} We examine candidate position, count, and spacing through order randomization, same-item scale refinement with balanced gold support and positions, and fixed-count candidate-spacing interventions. Wider ordinal gaps do not guarantee recovery of candidate utilization. Probability diagnostics further show that final decisions compress while average candidate probabilities remain broad.
  \item \textbf{Mitigating:} We show that targeted BA-LoRA post-training restores candidate utilization on eight supervised scales at both KEV sizes. Transfer to unseen ordinal sources is positive at 4B but heterogeneous at 0.8B, identifying compression as modifiable while separating in-domain recovery from general debiasing.
\end{itemize}

\section{Related Work}

\paragraph{Direct-decision models.}
Most language-model classifiers generate a label or score label tokens under a generative model \citep{holtzman-etal-2021-surface,robinson2023leveraging}.
Direct-decision models instead expose the candidate set as part of the input and return a distribution over candidates.
JEV supports structured choice, score, and \texttt{noul} decisions through the System One API \citep{typesafe2026api}.
The KEV family is an open reimplementation on Qwen3.5 backbones that follows the same interface \citep{qwen3.5,palmer2026kev,hume2026jev}.
Existing studies compare JEV with LLM judges in accuracy, cost, and latency, or use it as the first stage of an evaluation cascade \citep{rao2026jevrubric,li2026jevjudge}.
They do not systematically test whether its final decisions use the supplied ordinal scale.

\paragraph{Biases in model-based decisions.}
Fixed-option language decisions are sensitive to label priors, surface forms, option identifiers, prompt format, and candidate order \citep{zhao2021calibrate,holtzman-etal-2021-surface,lu-etal-2022-fantastically,fei-etal-2023-mitigating,zheng2024mcq,pezeshkpour-hruschka-2024-large}.
LLM judges also show position bias, verbosity bias, self-preference, and skewed score distributions \citep{zheng2023judging,wang-etal-2024-large-language-models-fair,saito2023verbosity,panickssery2024llm,stureborg2024inconsistent,li2025scoring}.
In particular, \citet{stureborg2024inconsistent} show that generative evaluators sparsely use a 1--100 scoring range, and \citet{rao2026jevrubric} report that JEV and LLM judges often assign lower rubric levels than human raters.
We study a related but distinct bias in direct decisions.
The shared property is not a fixed preference for low scores, Neutral, or one presentation position; it is the restricted use of model- and task-specific regions of an ordinal scale.

\section{Methods}

\subsection{Models}

We evaluate nine direct-decision models from three families.
JEV~1.13 is accessed through the hosted System One endpoint (reported version \texttt{typesafe/jev-1.13-20260917}); KEV-0.8B,\footnote{\url{https://huggingface.co/jaredpalmer/kev-0.8b}} KEV-4B,\footnote{\url{https://huggingface.co/jaredpalmer/kev-4b}} and KEV-9B\footnote{\url{https://huggingface.co/jaredpalmer/kev-9b}} are open JEV-like models initialized from Qwen3.5 checkpoints \citep{qwen3.5,palmer2026kev}.
We additionally evaluate StartLux Decision at 0.8B, 2B, 4B, 9B, and 27B parameters.\footnote{\url{https://github.com/StartLuxLabs/StartLux-Decision}}
Each model receives the text, question, and candidate list and returns candidate probabilities and an argmax decision.
We map outputs to semantic labels by candidate text, so comparisons use canonical labels rather than raw presentation indices.

\subsection{Datasets and Task Taxonomy}

The main panel contains 40 datasets with 5,000 items each, sampled without replacement and stratified by gold label using fixed seeds (Appendix~\ref{sec:appendix-datasets}).
We classify 36 datasets as \emph{ordinal}: their candidates form a shared graded scale and adjacent levels are semantically close.
We call this relation \emph{semantic coupling} as an operational ordinal/nominal distinction, not a continuous measurement.
The remaining four datasets are \emph{nominal}: SNIPS ($K=7$), Yahoo Answers ($K=10$), SCOTUS ($K=13$), and DBpedia ($K=14$) use unordered semantic categories.
Appendix~\ref{sec:appendix-datasets} lists the datasets and details sampling, candidate presentation, and label mapping; source-specific licenses and access terms are in Table~\ref{tab:dataset-licenses}.
Because gold distributions need not be uniform, we normalize utilization by the effective gold support and separately report distributional divergence.

\subsection{Measuring Scale-Utilization Bias}
\label{sec:metrics}

Let $p$ be the empirical distribution of argmax decisions and $q$ the gold-label distribution over $K$ candidates.
Let $H(r)$ denote the entropy of a label distribution $r$:
\[
  H(r)=-\sum_{j=1}^{K}r_j\log r_j.
\]
Then $\exp(H(r))$ represents the effective number of labels in $r$ \citep{hill1973diversity,jost2006entropy}.
We measure decision-level coverage with \textbf{argmax-label utilization} and the \textbf{absolute support deviation}:
\begin{equation}
  \label{eq:ueff}
  \begin{array}{rl}
    U_{\mathrm{arg}}&=\displaystyle\frac{\exp(H(p))}{K},\\[6pt]
    D_R&=\displaystyle100\left|\frac{\exp(H(p))}{\exp(H(q))}-1\right|.
  \end{array}
\end{equation}
$U_{\mathrm{arg}}$ compares decisions with the nominal space, whereas $D_R$ measures their effective-support deviation from the gold labels in percentage points.
Lower $D_R$ is better; zero denotes equal effective support, not equal distributional shape. Total variation distance (TVD) measures the latter.
We call a shortfall in predicted effective support, $\exp(H(p))<\exp(H(q))$, \emph{candidate-space compression}, without imposing a binary threshold.
Because $D_R$ also counts support expansion, we interpret its magnitude together with this direction and TVD.
We call a candidate that attracts a disproportionate share of final decisions an \emph{anchor}.
Accuracy alone does not establish compression.

For the mean label-space probability vector $\bar\pi$, soft utilization is
\begin{equation}
  \label{eq:soft-utilization}
  U_{\mathrm{soft}}=\frac{\exp(H(\bar\pi))}{K}.
\end{equation}
We additionally report normalized entropy, top-1 margin, normalized mean absolute error (nMAE) \citep{mathew2026llmsgrade}, and quadratic weighted kappa (QWK) \citep{cohen1968weighted,fleisscohen1973equivalence}.
These metrics distinguish decision-level compression from probability concentration and ordinal error.
Appendix~\ref{sec:appendix-metrics} provides complete definitions; all metrics are computed per dataset and macro-averaged.

\subsection{Candidate-Order Intervention}

We test whether fixed presentation positions explain compression on six ordinal and three nominal datasets.
For each of their 5,000 items, we generate two candidate permutations that differ from each other and from the original order.
We compare the original order with the mean of the two randomized orders using accuracy, TVD, and $D_R$.
Each metric is computed separately for each dataset and order before averaging over the two orders and then across datasets.
Pairwise label agreement and the TVD between the two label-space probability vectors measure item-level position sensitivity; Appendix~\ref{sec:appendix-datasets} lists the selected datasets and implementation details.

\subsection{Controlled Scale Refinement}

Cross-dataset comparisons can confound candidate count with task difficulty and label structure.
To study candidate count, we vary $K\in\{2,\ldots,14\}$ on the same 1,400 items from three continuous-score sources: RealToxicity Continuation \citep{gehman-etal-2020-realtoxicityprompts}, Civil Comments Toxicity \citep{borkan2019nuanced}, and STS Benchmark Similarity \citep{cer-etal-2017-semeval}.
Equal-width bins preserve the natural score scale, whereas equal-frequency bins keep the gold levels close to balanced.
Candidate order is randomized and the gold position is balanced over the $K$ positions in every condition.
As a nominal reference, 1,400 DBpedia items \citep{zhang2015character} receive a random $K$-class subset containing the gold class; we compute $D_R$ within these subsets and use DBpedia as a ceiling reference rather than a difficulty-matched control.
Appendix~\ref{sec:appendix-construction} specifies the binning, interval boundaries, candidate text, and subset aggregation.

\subsection{Fixed-Count Candidate-Distance Intervention}
\label{sec:distance-method}

Scale refinement changes both candidate count and the distinctions between neighboring levels.
We test whether difficulty distinguishing nearby ordinal levels explains compression by changing candidate spacing at fixed $K$.
We use Word Concreteness, WMT20 Translation Quality, Measuring Hate Speech, and Stanford Politeness, retaining their original questions and ordinal labels.
After removing repeated contexts and balancing original gold levels, the four sources contribute 4,968, 4,968, 4,992, and 4,914 unique items, respectively.
Using the fixed seed 20261007, each item receives seven candidate sets, yielding 138,894 questions for each KEV and StartLux checkpoint.
JEV~1.13 is evaluated on the three-source common panel excluding Word Concreteness (104,118 questions).

For each source, we partition its original scale into equal-sized candidate sets and show the set containing the item's gold label.
$K$ is fixed at three for Word Concreteness, WMT20, and Politeness, and four for Hate Speech.
Spacing is the mean pairwise distance between original ordinal ranks, normalized by the original scale's rank range.
We enumerate partitions, retain the lowest-dispersion partition for each attainable mean, and select seven distinct means nearest to equally spaced targets between the minimum and maximum.
The smallest-spacing condition groups consecutive levels, whereas the largest uses interleaved levels spanning the scale.
The text, question, correct label, and displayed gold position remain fixed across an item's versions.
Original gold labels, displayed gold positions, and relative gold ranks are balanced in every condition; each candidate in every ordered set serves equally often as gold.
An item's relative gold rank may change across conditions, and spacing increases at the condition-average level.
We compute $D_R$ and TVD over relative ordinal ranks among the displayed candidates, separately for each dataset and distance condition.
This intervention measures ordinal-rank spacing rather than a general semantic-distance score.
Appendix~\ref{sec:appendix-distance-construction} gives the distance formula, exact sample balance, seven spacing values, and candidate examples.

\subsection{Bias-Aware Post-Training}
\label{sec:balora-method}

We test whether candidate-space compression can be modified after pretraining by adapting KEV-0.8B and KEV-4B with BA-LoRA \citep{chang2026balora}.
The fixed post-training set contains 32,000 examples, with 4,000 examples from each of eight sources that exhibit strong ordinal compression.
The source ID and normalized-text hash of every post-training example are disjoint from the frozen 40-dataset evaluation panel.
We report dataset-level macro-averages separately for the eight training-source datasets, the other 28 ordinal datasets, all 36 ordinal datasets, and the four nominal datasets.
Appendix~\ref{sec:appendix-posttraining} lists the sources and complete training configuration; Appendix~\ref{sec:appendix-posttraining-results} reports the full numerical results and per-dataset illustrations.

\section{Experiments}

\subsection{Main Results}
\label{sec:broad}

\begin{table*}[t]
  \centering
  \footnotesize
  \setlength{\tabcolsep}{4pt}
  \begin{tabular}{lccccccccccc}
    \toprule
    & \multicolumn{8}{c}{\textbf{Ordinal (36 datasets)}} & \multicolumn{3}{c}{\textbf{Nominal (4 datasets)}} \\
    \cmidrule(lr){2-9} \cmidrule(lr){10-12}
    \textbf{Model} & Acc $\uparrow$ & $U_{\mathrm{arg}}$ & $D_R$ $\downarrow$ & TVD $\downarrow$ & nMAE $\downarrow$ & QWK $\uparrow$ & $U_{\mathrm{soft}}$ & Margin & Acc $\uparrow$ & $D_R$ $\downarrow$ & TVD $\downarrow$ \\
    \midrule
    JEV~1.13 & \textbf{37.67} & 72.97 & \textbf{25.92} & \textbf{29.82} & 20.78 & \textbf{0.593} & 89.10 & 36.16 & 84.39 & 1.47 & 6.04 \\
    \midrule
    KEV-0.8B & 28.80 & 66.87 & 31.11 & 37.43 & 30.47 & 0.370 & 97.10 & 17.49 & 69.63 & 13.37 & 18.58 \\
    KEV-4B   & 33.37 & 65.32 & 32.80 & 37.48 & 24.70 & 0.455 & 93.61 & 20.00 & 79.74 & 3.81 & 9.89 \\
    KEV-9B   & 35.02 & 70.46 & 29.28 & 33.21 & 22.17 & 0.523 & 94.63 & 18.46 & 81.80 & 3.31 & 6.94 \\
    \midrule
    StartLux-0.8B & 31.75 & 58.91 & 39.53 & 42.98 & 26.84 & 0.391 & 95.31 & 11.03 & 77.14 & 3.76 & 10.87 \\
    StartLux-2B   & 33.20 & 62.70 & 35.69 & 39.96 & 24.20 & 0.459 & 95.24 & 13.12 & 81.92 & 1.16 & 6.60 \\
    StartLux-4B   & 35.00 & 66.08 & 32.14 & 36.55 & 22.13 & 0.528 & 90.67 & 18.05 & 84.00 & 1.23 & \textbf{4.70} \\
    StartLux-9B   & 36.89 & 67.58 & 30.40 & 34.41 & 20.98 & 0.570 & 91.79 & 19.73 & 84.30 & 1.22 & 5.46 \\
    StartLux-27B  & 37.24 & 68.79 & 29.33 & 33.36 & \textbf{20.10} & 0.579 & 89.63 & 19.10 & \textbf{84.64} & \textbf{0.99} & 5.23 \\
    \bottomrule
  \end{tabular}
  \caption{\label{tab:main}
    \textbf{Main-panel results.} Dataset macro-averages. Units: $D_R$/Margin in pp; QWK unscaled; others in \%. Bold marks column-best values where arrows specify direction.
  }
\end{table*}

\paragraph{Candidate-space compression is widespread but structure-dependent.}
Across the 36 ordinal datasets, mean $D_R$ ranges from 25.9 to 39.5 points and TVD from 29.8 to 43.0\% (Table~\ref{tab:main}).
Each model has smaller predicted than gold effective support on 35--36 datasets (Appendix~\ref{sec:appendix-main-results}).
$D_R$ exceeds 30 points on 14--21 datasets per model, showing that the pattern is not driven by a few extreme cases.
On the balanced 14-level RealToxicity task, JEV~1.13 assigns 66.2\% of predictions to Level 01, yielding $D_R\approx75$ points.
By contrast, the four nominal tasks have mean $D_R=1.0$--13.4 points despite offering 7--14 candidates.
Their TVD is also lower, at 4.7--18.6\%.
An exception is SCOTUS, where KEV-0.8B assigns 55\% of predictions to ``Miscellaneous'' ($D_R\approx49$ points).
Nominal tasks generally use their label spaces more completely, with dataset-level variation reported in Appendix~\ref{sec:appendix-main-results}.

\paragraph{Candidate count is associated with compression across domains.}
Across ordinal datasets, $D_R$ increases with $K$ for the JEV/KEV models (Spearman $\rho=0.74$--$0.82$).
This association spans multiple domains, although $K$ covaries with other dataset properties.
Section~\ref{sec:kscaling} therefore tests scale refinement on the same items.

\paragraph{Compression has no universal direction.}
The modal prediction can fall in the lower, middle, or upper part of an ordinal scale, depending on the model and dataset.
JEV's mode is in these three regions on 21, 9, and 6 datasets, respectively; KEV-0.8B's corresponding counts are 15, 9, and 12.
The shared pattern is reduced effective support, not a universal preference for one score region (Appendix~\ref{sec:appendix-main-results}).

\paragraph{Larger models remain biased despite higher accuracy.}
KEV accuracy rises from 28.8\% at 0.8B to 33.4\% at 4B and 35.0\% at 9B, whereas $D_R$ changes from 31.1 to 32.8 to 29.3 points.
KEV-4B is therefore more accurate than KEV-0.8B while showing greater compression.
StartLux improves more steadily, but its 27B checkpoint still has $D_R=29.3$ points despite reaching 37.2\% accuracy (Table~\ref{tab:main}).

\subsection{Analysis}
\label{sec:analysis}

\subsubsection{Randomized Candidate Positions}
\label{sec:position}

\begin{table*}[t]
  \centering
  \footnotesize
  \setlength{\tabcolsep}{4pt}
  \renewcommand{\arraystretch}{0.94}
  \begin{tabular}{llcccccccc}
    \toprule
    & & \multicolumn{2}{c}{\textbf{Acc} $\uparrow$} & \multicolumn{2}{c}{\textbf{TVD} $\downarrow$} & \multicolumn{2}{c}{{\boldmath$D_R$} $\downarrow$} & \multicolumn{2}{c}{\textbf{Rand.\ 1 vs.\ 2}} \\
    \cmidrule(lr){3-4} \cmidrule(lr){5-6} \cmidrule(lr){7-8} \cmidrule(lr){9-10}
    & \textbf{Model} & Orig. & Rand. & Orig. & Rand. & Orig. & Rand. & Agree. $\uparrow$ & Shift $\downarrow$ \\
    \midrule
    \multirow{9}{*}{\rotatebox[origin=c]{90}{\textbf{Ordinal}}}
    & JEV~1.13 & 26.99 & 26.88 & \textbf{35.61} & 32.25 & \textbf{34.99} & 29.65 & 74.43 & 9.42 \\
    \cmidrule(lr){2-10}
    & KEV-0.8B & 22.19 & 21.68 & 41.87 & 31.41 & 38.94 & 26.28 & 58.58 & 9.65 \\
    & KEV-4B   & 24.56 & 24.92 & 45.73 & \textbf{29.30} & 44.70 & \textbf{26.06} & 58.14 & 10.13 \\
    & KEV-9B   & 25.79 & 25.30 & 45.39 & 35.05 & 43.01 & 32.31 & 59.94 & 9.47 \\
    \cmidrule(lr){2-10}
    & StartLux-0.8B & 25.64 & 25.87 & 50.94 & 45.40 & 52.46 & 46.03 & 76.30 & \textbf{5.33} \\
    & StartLux-2B   & 26.34 & 26.40 & 47.21 & 35.84 & 46.03 & 32.19 & 65.99 & 7.00 \\
    & StartLux-4B   & 26.65 & 27.59 & 43.06 & 31.76 & 43.38 & 29.26 & 69.37 & 7.59 \\
    & StartLux-9B   & 28.36 & \textbf{28.74} & 42.76 & 36.50 & 42.12 & 34.34 & \textbf{77.99} & 6.06 \\
    & StartLux-27B  & \textbf{28.67} & 28.54 & 43.48 & 36.91 & 43.47 & 35.42 & 74.91 & 7.16 \\
    \midrule
    \multirow{9}{*}{\rotatebox[origin=c]{90}{\textbf{Nominal}}}
    & JEV~1.13 & 89.43 & 89.55 & 4.13 & 3.54 & 0.94 & 0.66 & 98.21 & 2.36 \\
    \cmidrule(lr){2-10}
    & KEV-0.8B & 83.72 & 83.36 & 5.85 & 5.46 & 1.50 & 1.37 & 95.59 & 4.71 \\
    & KEV-4B   & 88.07 & 87.52 & 4.53 & 4.85 & 1.30 & 1.17 & 96.33 & 4.74 \\
    & KEV-9B   & 88.50 & 88.06 & 4.40 & 4.73 & 1.08 & 1.19 & 96.81 & 5.13 \\
    \cmidrule(lr){2-10}
    & StartLux-0.8B & 87.81 & 87.83 & 4.38 & 4.05 & 1.11 & 0.92 & 98.21 & 3.11 \\
    & StartLux-2B   & 88.41 & 88.80 & 4.99 & 4.50 & 1.22 & 0.99 & 97.87 & 3.16 \\
    & StartLux-4B   & 89.79 & 89.90 & \textbf{3.19} & \textbf{2.93} & \textbf{0.47} & \textbf{0.45} & 98.25 & 2.74 \\
    & StartLux-9B   & \textbf{90.36} & \textbf{90.57} & 3.69 & 3.12 & 0.74 & 0.58 & \textbf{98.91} & \textbf{1.67} \\
    & StartLux-27B  & 89.87 & 90.11 & 3.57 & 3.25 & 0.71 & 0.55 & 98.53 & 2.39 \\
    \bottomrule
  \end{tabular}
  \caption{\label{tab:position}
    \textbf{Candidate-order results} (six ordinal, three nominal datasets). Rand.: metrics averaged over two orders, then datasets. Agree./Shift: canonical-label agreement/probability TVD between orders. Units: $D_R$/Shift in pp; others in \%. Bold marks column-best models within each task type.
  }
\end{table*}

\paragraph{Candidate order changes individual predictions with little change in macro accuracy.}
The two randomized orders agree on 58.1--78.0\% of ordinal items, compared with at least 95.6\% of nominal items (Table~\ref{tab:position}).
Probability shifts are also larger on ordinal tasks: 5.3--10.1 TVD points, compared with 1.7--5.1 points on nominal tasks.
Yet randomization changes macro accuracy by at most 0.95 points for every model and task type.
Correct-to-incorrect and incorrect-to-correct changes can offset in aggregate, while switches between incorrect labels leave accuracy unchanged.
Stable macro accuracy therefore conceals order-sensitive decisions and probabilities (Appendix~\ref{sec:appendix-order-results}).

\paragraph{Randomization weakens ordinal-scale bias but leaves a large residual.}
Across the six ordinal datasets, randomization lowers every model's mean $D_R$, from 35.0--52.5 points to 26.1--46.0 points.
TVD also falls by 3.4--16.4 points, while randomized nominal $D_R$ remains within 0.4--1.4 points.
Thus, candidate order affects individual decisions, but randomization does not eliminate aggregate compression.

\paragraph{Order sensitivity does not account for the full distributional bias.}
Random permutations distribute a fixed positional preference across canonical labels rather than preserving one label in a favored slot.
A purely first-, middle-, or last-position mechanism would therefore be weakened by randomization.
Nevertheless, the ordinal predictions remain concentrated even as many individual answers change.
Order sensitivity and aggregate scale use are distinct: stabilizing the presentation order would not itself establish a gold-aligned prediction distribution.
Appendix~\ref{sec:appendix-order-results} provides model-specific residuals and aggregation details.

\subsubsection{Candidate Count \texorpdfstring{$K$}{K}}
\label{sec:kscaling}

\begin{figure*}[t]
  \centering
  \includegraphics[width=\textwidth]{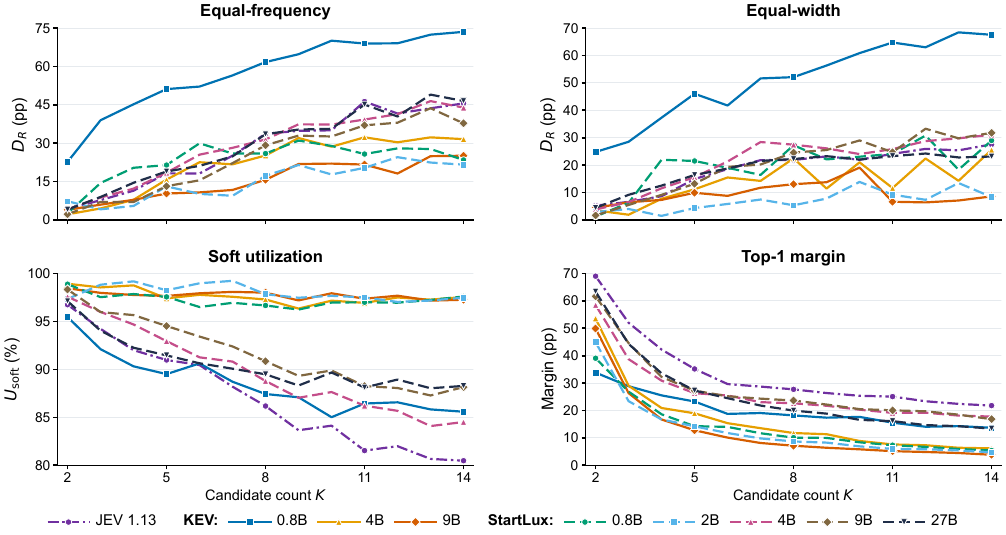}
  \caption{\label{fig:kcurve}
    \textbf{Same-item scale refinement.} Macro-averages over three ordinal datasets (1,400 fixed items each). Upper row: equal-frequency and equal-width $D_R$. Lower row: equal-frequency soft utilization and top-1 margin. Units: $D_R$/margin in pp; utilization in \%. Linear, panel-specific y-ranges.
  }
\end{figure*}

\paragraph{Increasing $K$ strengthens compression on the same items.}
With text and source scores fixed, equal-frequency binning keeps gold effective-label coverage near-balanced: 99.8\% at $K=2$ and 97.0\% at $K=14$.
Over these endpoints, JEV~1.13's $D_R$ rises from 4.1 to 45.6 points (Figure~\ref{fig:kcurve}, upper left).
The KEV endpoints are 22.7 to 73.5 points at 0.8B, 2.1 to 31.6 at 4B, and 3.9 to 25.1 at 9B.
All nine models show endpoint increases of 14.3--50.9 points, although intermediate curves need not be monotonic (Appendix Table~\ref{tab:kcurve-endpoints}).
The effective decision space therefore grows more slowly than the available ordinal scale.
A decrease in exact-match accuracy alone would not establish this effect.
$D_R$ instead tests whether predicted effective support keeps pace with the finer gold-label distribution.
Equal-width binning also increases endpoint $D_R$ for every model; construction-specific differences appear in Appendix~\ref{sec:appendix-k-results}.
For KEV-9B, the equal-width endpoints are 4.4 and 8.6 points, compared with 3.9 and 25.1 under equal-frequency binning.
The direction is shared, but its magnitude depends on how the ordinal scale is constructed.

\paragraph{Decision concentration coexists with broad average probability coverage.}
For KEV-4B and KEV-9B, $U_{\mathrm{soft}}$ remains at least 96\% at every $K$ under both binnings.
Their mean top-1 margins nevertheless fall from 50--54 points at $K=2$ to 4--6 points at $K=14$ (Figure~\ref{fig:kcurve}, lower right).
Thus, broad average probability coverage does not prevent increasingly concentrated final decisions.
JEV also loses soft coverage, while its margin falls from 69.0 to 21.8 points under equal-frequency binning.
For JEV and KEV, endpoint nMAE stays comparatively stable or improves despite increasing $D_R$.
The growing distributional bias is therefore not captured by ordinal error alone, just as high average probability coverage does not establish broad decision utilization.
Appendix~\ref{sec:appendix-k-results} reports the other models' probability trends and accompanying ordinal-error metrics.

\paragraph{Candidate count alone does not determine scale utilization.}
On DBpedia, JEV~1.13 maintains at least 98.9\% accuracy and subset-weighted $D_R<0.4$ points throughout $K=2$--14.
This ceiling reference shows that many candidates do not inevitably produce compression, unlike refinement of the tested ordinal scales.
The next intervention varies ordinal spacing at fixed $K$; DBpedia diagnostics are reported in Appendix~\ref{sec:appendix-k-results}.

\subsubsection{Fixed-Count Candidate Distance}
\label{sec:distance}

\begin{figure*}[t]
  \centering
  \begin{minipage}[t]{\columnwidth}
    \vspace{0pt}
    \centering
    \includegraphics[width=\linewidth]{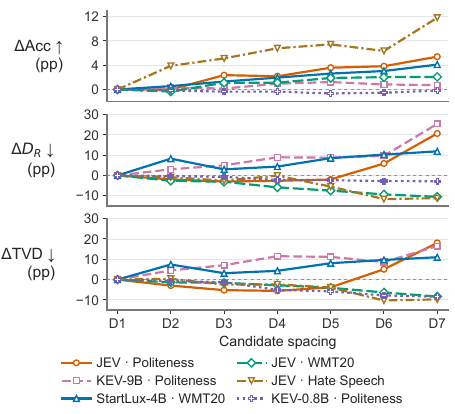}
    \caption{\label{fig:candidate-distance}
      \textbf{Fixed-count spacing.} Changes from D1 for six representative pairs, evaluated on fixed items. Arrows mark preferred directions. Full results: Appendix~\ref{sec:appendix-distance-results}.
    }
  \end{minipage}\hfill
  \begin{minipage}[t]{\columnwidth}
    \vspace{0pt}
    \input{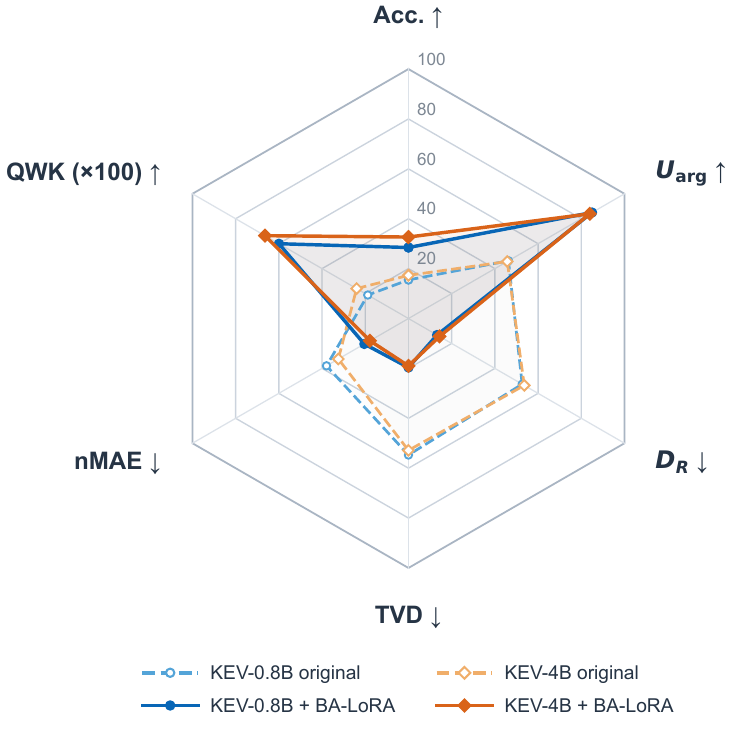}
  \end{minipage}
\end{figure*}

\paragraph{Greater candidate spacing can improve correctness without restoring utilization.}
Items, correct labels, displayed gold positions, and $K$ remain fixed (Section~\ref{sec:distance-method}).
Across the three shared sources, JEV's macro accuracy rises from 36.3\% to 42.7\%, while $D_R$ changes from 16.6 to 16.3 points and TVD stays near 26.5\%.
This average masks decreasing bias on WMT20 and Hate Speech and increasing bias on Politeness (Figure~\ref{fig:candidate-distance}).
On WMT20, StartLux-4B's accuracy rises from 31.9\% to 36.0\%, but $D_R$ increases from 15.0 to 26.8 points and TVD from 27.7\% to 38.7\%.
Accuracy improves at every step, yet both bias metrics worsen at five of six transitions.

\paragraph{Wider spacing can strengthen an anchor while accuracy improves.}
On Politeness, JEV's accuracy rises from 38.2\% to 43.6\%, but $D_R$ increases from 6.1 to 26.8 points and TVD from 16.9\% to 34.9\%.
The middle-ranked candidate's prediction share rises from 22.3\% to 68.3\%, against a constant one-third gold share.
Bias first falls and then rises, rather than worsening monotonically.
KEV-9B shows the same endpoint dissociation (Figure~\ref{fig:candidate-distance}).
Conversely, KEV-0.8B reaches $D_R=0.1$ at D7 with 34.8\% accuracy, near the 33.3\% chance level.
Thus, neither greater accuracy nor matching effective support guarantees the other.

\paragraph{Spacing and scale refinement produce different responses.}
Increasing $K$ consistently increases endpoint compression, whereas increasing spacing at fixed $K$ has model- and task-dependent effects.
Wider gaps change the competing alternatives, but all candidates still lie on the same evaluative axis.
The intervention therefore separates local ordinal spacing from the broader distinction between graded scales and unordered categories.
Correctness and aggregate scale use therefore require separate assessment: wider ordinal gaps are not a general remedy for compression.
Complete results, including Word Concreteness, appear in Appendix~\ref{sec:appendix-distance-results}.

\subsection{BA-LoRA Post-Training}
\label{sec:posttraining}

\paragraph{Targeted post-training restores candidate-space utilization at both KEV scales.}
On disjoint held-out items from the eight training sources, BA-LoRA lowers $D_R$ from 52.6 to 13.2 points for KEV-0.8B.
For KEV-4B, $D_R$ falls from 53.6 to 14.4 points (Figure~\ref{fig:ba-lora-radar}).
Accuracy increases by 13.0 and 15.3 points, respectively; every targeted source improves in accuracy, $D_R$, and TVD for both sizes.
TVD falls from 54.7\% to 19.7\% for KEV-0.8B and from 52.9\% to 19.0\% for KEV-4B, so the recovery also improves label-frequency alignment.
The candidate labels and evaluation items are unchanged, so recovery does not require coarsening the scale or redefining the task.

\paragraph{Broader utilization accompanies better ordinal decisions.}
Across all 36 ordinal datasets, accuracy rises from 28.8\% to 32.3\% for KEV-0.8B and from 33.4\% to 37.9\% for KEV-4B.
Both models also improve in $D_R$, nMAE, and QWK (Appendix Table~\ref{tab:balora}).
For KEV-0.8B, nMAE falls from 30.5\% to 24.0\% and QWK rises from 0.370 to 0.488.
KEV-4B shows the same pattern, with nMAE falling from 24.7\% to 19.3\% and QWK rising from 0.455 to 0.606.
Utilization gains therefore accompany improvements in correctness, not just a broader prediction distribution.
On the eight training sources, $U_{\mathrm{arg}}$ rises by 38.7 and 38.2 points, whereas $U_{\mathrm{soft}}$ increases by 0.9 and 6.3 points, respectively.
The larger change is in final label selection rather than the effective coverage of the average probability vector.

\paragraph{Transfer to unseen sources differs between the two models.}
On the other 28 ordinal datasets, KEV-4B reduces $D_R$ from 26.8 to 21.3 points and TVD from 33.1\% to 27.6\%.
KEV-0.8B instead increases $D_R$ from 25.0 to 29.9 points and TVD from 32.5\% to 35.1\%.
Macro-averaged over the four nominal datasets, accuracy and TVD improve for both models, whereas $D_R$ improves only for KEV-0.8B.
For KEV-4B, mean nominal $D_R$ rises from 3.8 to 4.5 points.
On SCOTUS, KEV-0.8B accuracy rises from 27.4\% to 42.1\% and TVD falls from 56.8\% to 32.2\%.
On Yahoo Answers, KEV-4B accuracy changes from 70.2\% to 70.3\%, while TVD rises from 11.4\% to 13.8\%.
The contrast separates recovery on supervised scales from transfer to other tasks.
Appendix~\ref{sec:appendix-posttraining-results} provides the full numerical results (Table~\ref{tab:balora}) and selected per-dataset illustrations; Appendix~\ref{sec:appendix-posttraining} gives the training configuration.

\section{Conclusion}

Across nine direct-decision models and 40 datasets, we identify ordinal scale-utilization bias: decisions concentrate on model- and task-specific anchors rather than a universally favored label.
Nominal tasks retain broader effective support on average despite their often large candidate sets.
The central issue is therefore how models use graded scales.

Controlled interventions sharpen this characterization: randomizing positions weakens compression, yet finer ordinal scales strengthen it on fixed items with near-balanced gold support and controlled positions.
Wider gaps at fixed candidate count do not reliably restore utilization, even when accuracy improves.
For most models, broad average probabilities coexist with concentrated argmax decisions and narrowing margins, separating probability coverage from actual scale use.

Taken together, the evidence identifies a strong dependence on ordinal structure and greater compression as the candidate scale is refined.
Neither presentation position nor local ordinal spacing alone sufficiently explains this pattern.
Larger models improve accuracy without consistently restoring scale utilization, but this bias can be effectively mitigated through targeted training.
Scale utilization is therefore an actionable reliability target alongside prediction accuracy.

\clearpage
\section*{Limitations}

\paragraph{Scope of the evidence.}
Semantic coupling is operationalized through a binary ordinal/nominal taxonomy, so our account of why ordinal tasks are more compressed rests on task type rather than on a graded measure of coupling.
The same-item scaling result covers three ordinal sources.
Its nominal reference is DBpedia, on which all models are near ceiling.
This reference shows that more candidates alone need not produce compression, but it does not provide a difficulty-matched nominal comparison.
The position result is based on two random permutations per item, and all results are point estimates.
$D_R$ compares the effective support sizes of the predicted and gold distributions; $D_R=0$ does not imply that the two distributions assign mass to the same labels, which is why we report TVD separately.

\paragraph{Models and data.}
We study one hosted system (\texttt{typesafe/jev-1.13-20260917}) and two open families (KEV and StartLux), so the findings may not transfer to other direct-decision models.
SST-5, DBpedia-14, and IMDb are listed among the KEV training sources.
We did not audit item-level overlap with our samples.
KEV results on these three datasets, including the DBpedia reference, may therefore partly reflect training exposure.

\bibliography{custom}

@inproceedings{frank-hall-2001-ordinal,
  title     = {A Simple Approach to Ordinal Classification},
  author    = {Frank, Eibe and Hall, Mark},
  booktitle = {Machine Learning: {ECML} 2001},
  series    = {Lecture Notes in Computer Science},
  volume    = {2167},
  pages     = {145--156},
  year      = {2001},
  publisher = {Springer},
  doi       = {10.1007/3-540-44795-4_13},
  url       = {https://doi.org/10.1007/3-540-44795-4_13}
}

@article{gutierrez-etal-2016-ordinal,
  title   = {Ordinal Regression Methods: Survey and Experimental Study},
  author  = {Guti{\'e}rrez, Pedro Antonio and P{\'e}rez-Ortiz, Mar{\'i}a and S{\'a}nchez-Monedero, Javier and Fern{\'a}ndez-Navarro, Francisco and Herv{\'a}s-Mart{\'i}nez, C{\'e}sar},
  journal = {IEEE Transactions on Knowledge and Data Engineering},
  volume  = {28},
  number  = {1},
  pages   = {127--146},
  year    = {2016},
  doi     = {10.1109/TKDE.2015.2457911},
  url     = {https://doi.org/10.1109/TKDE.2015.2457911}
}

@inproceedings{liu-etal-2023-g,
  title     = {{G}-Eval: {NLG} Evaluation Using {GPT}-4 with Better Human Alignment},
  author    = {Liu, Yang and Iter, Dan and Xu, Yichong and Wang, Shuohang and Xu, Ruochen and Zhu, Chenguang},
  booktitle = {Proceedings of the 2023 Conference on Empirical Methods in Natural Language Processing},
  month     = dec,
  year      = {2023},
  address   = {Singapore},
  publisher = {Association for Computational Linguistics},
  pages     = {2511--2522},
  doi       = {10.18653/v1/2023.emnlp-main.153},
  url       = {https://aclanthology.org/2023.emnlp-main.153/}
}

@inproceedings{fei-etal-2023-mitigating,
  title     = {Mitigating Label Biases for In-Context Learning},
  author    = {Fei, Yu and Hou, Yifan and Chen, Zeming and Bosselut, Antoine},
  booktitle = {Proceedings of the 61st Annual Meeting of the Association for Computational Linguistics (Volume 1: Long Papers)},
  month     = jul,
  year      = {2023},
  address   = {Toronto, Canada},
  publisher = {Association for Computational Linguistics},
  pages     = {14014--14031},
  doi       = {10.18653/v1/2023.acl-long.783},
  url       = {https://aclanthology.org/2023.acl-long.783/}
}

@article{cohen1968weighted,
  title   = {Weighted Kappa: Nominal Scale Agreement with Provision for Scaled Disagreement or Partial Credit},
  author  = {Cohen, Jacob},
  journal = {Psychological Bulletin},
  volume  = {70},
  number  = {4},
  pages   = {213--220},
  year    = {1968},
  doi     = {10.1037/h0026256},
  url     = {https://doi.org/10.1037/h0026256}
}

@article{fleisscohen1973equivalence,
  title   = {The Equivalence of Weighted Kappa and the Intraclass Correlation Coefficient as Measures of Reliability},
  author  = {Fleiss, Joseph L. and Cohen, Jacob},
  journal = {Educational and Psychological Measurement},
  volume  = {33},
  number  = {3},
  pages   = {613--619},
  year    = {1973},
  doi     = {10.1177/001316447303300309},
  url     = {https://doi.org/10.1177/001316447303300309}
}

@misc{mathew2026llmsgrade,
  title         = {{LLMs} Do Not Grade Essays Like Humans},
  author        = {Mathew, Jerin George and Taher, Sumayya and Kundu, Anindita and Barbosa, Denilson},
  year          = {2026},
  eprint        = {2603.23714},
  archivePrefix = {arXiv},
  primaryClass  = {cs.AI},
  url           = {https://arxiv.org/abs/2603.23714}
}

@article{hill1973diversity,
  title   = {Diversity and Evenness: A Unifying Notation and Its Consequences},
  author  = {Hill, M. O.},
  journal = {Ecology},
  volume  = {54},
  number  = {2},
  pages   = {427--432},
  year    = {1973},
  doi     = {10.2307/1934352},
  url     = {https://doi.org/10.2307/1934352}
}

@article{jost2006entropy,
  title   = {Entropy and Diversity},
  author  = {Jost, Lou},
  journal = {Oikos},
  volume  = {113},
  number  = {2},
  pages   = {363--375},
  year    = {2006},
  doi     = {10.1111/j.2006.0030-1299.14714.x},
  url     = {https://doi.org/10.1111/j.2006.0030-1299.14714.x}
}

@inproceedings{zheng2024mcq,
  title     = {Large Language Models Are Not Robust Multiple Choice Selectors},
  author    = {Zheng, Chujie and Zhou, Hao and Meng, Fandong and Zhou, Jie and Huang, Minlie},
  booktitle = {The Twelfth International Conference on Learning Representations},
  year      = {2024},
  url       = {https://openreview.net/forum?id=shr9PXz7T0}
}

@inproceedings{pezeshkpour-hruschka-2024-large,
    title = "Large Language Models Sensitivity to The Order of Options in Multiple-Choice Questions",
    author = "Pezeshkpour, Pouya  and
      Hruschka, Estevam",
    editor = "Duh, Kevin  and
      Gomez, Helena  and
      Bethard, Steven",
    booktitle = "Findings of the Association for Computational Linguistics: NAACL 2024",
    month = jun,
    year = "2024",
    address = "Mexico City, Mexico",
    publisher = "Association for Computational Linguistics",
    url = "https://aclanthology.org/2024.findings-naacl.130/",
    doi = "10.18653/v1/2024.findings-naacl.130",
    pages = "2006--2017"
}

@inproceedings{wang-etal-2024-large-language-models-fair,
    title = "Large Language Models are not Fair Evaluators",
    author = "Wang, Peiyi  and Li, Lei  and Chen, Liang  and Cai, Zefan  and
      Zhu, Dawei  and Lin, Binghuai  and Cao, Yunbo  and Kong, Lingpeng  and
      Liu, Qi  and Liu, Tianyu  and Sui, Zhifang",
    editor = "Ku, Lun-Wei  and Martins, Andre  and Srikumar, Vivek",
    booktitle = "Proceedings of the 62nd Annual Meeting of the Association for Computational Linguistics (Volume 1: Long Papers)",
    month = aug,
    year = "2024",
    address = "Bangkok, Thailand",
    publisher = "Association for Computational Linguistics",
    url = "https://aclanthology.org/2024.acl-long.511/",
    doi = "10.18653/v1/2024.acl-long.511",
    pages = "9440--9450"
}

@inproceedings{zheng2023judging,
  title     = {Judging {LLM}-as-a-Judge with {MT}-Bench and Chatbot Arena},
  author    = {Zheng, Lianmin and Chiang, Wei-Lin and Sheng, Ying and Zhuang, Siyuan and Wu, Zhanghao and Zhuang, Yonghao and Lin, Zi and Li, Zhuohan and Li, Dacheng and Xing, Eric P. and Zhang, Hao and Gonzalez, Joseph E. and Stoica, Ion},
  booktitle = {Advances in Neural Information Processing Systems},
  volume    = {36},
  pages     = {46595--46623},
  year      = {2023},
  note      = {Datasets and Benchmarks Track},
  url       = {https://proceedings.neurips.cc/paper_files/paper/2023/hash/91f18a1287b398d378ef22505bf41832-Abstract-Datasets_and_Benchmarks.html}
}

@inproceedings{zhao2021calibrate,
  title     = {Calibrate Before Use: Improving Few-shot Performance of Language Models},
  author    = {Zhao, Zihao and Wallace, Eric and Feng, Shi and Klein, Dan and Singh, Sameer},
  booktitle = {Proceedings of the 38th International Conference on Machine Learning},
  series    = {Proceedings of Machine Learning Research},
  volume    = {139},
  pages     = {12697--12706},
  year      = {2021},
  publisher = {PMLR},
  url       = {https://proceedings.mlr.press/v139/zhao21c.html}
}

@inproceedings{robinson2023leveraging,
  title     = {Leveraging Large Language Models for Multiple Choice Question Answering},
  author    = {Robinson, Joshua and Rytting, Christopher Michael and Wingate, David},
  booktitle = {The Eleventh International Conference on Learning Representations},
  year      = {2023},
  url       = {https://arxiv.org/abs/2210.12353v3}
}

@inproceedings{holtzman-etal-2021-surface,
    title = "Surface Form Competition: Why the Highest Probability Answer Isn{'}t Always Right",
    author = "Holtzman, Ari  and West, Peter  and Shwartz, Vered  and Choi, Yejin  and Zettlemoyer, Luke",
    editor = "Moens, Marie-Francine  and Huang, Xuanjing  and Specia, Lucia  and Yih, Scott Wen-tau",
    booktitle = "Proceedings of the 2021 Conference on Empirical Methods in Natural Language Processing",
    month = nov,
    year = "2021",
    address = "Online and Punta Cana, Dominican Republic",
    publisher = "Association for Computational Linguistics",
    url = "https://aclanthology.org/2021.emnlp-main.564/",
    doi = "10.18653/v1/2021.emnlp-main.564",
    pages = "7038--7051"
}

@inproceedings{lu-etal-2022-fantastically,
    title = "Fantastically Ordered Prompts and Where to Find Them: Overcoming Few-Shot Prompt Order Sensitivity",
    author = "Lu, Yao  and Bartolo, Max  and Moore, Alastair  and Riedel, Sebastian  and Stenetorp, Pontus",
    editor = "Muresan, Smaranda  and Nakov, Preslav  and Villavicencio, Aline",
    booktitle = "Proceedings of the 60th Annual Meeting of the Association for Computational Linguistics (Volume 1: Long Papers)",
    month = may,
    year = "2022",
    address = "Dublin, Ireland",
    publisher = "Association for Computational Linguistics",
    url = "https://aclanthology.org/2022.acl-long.556/",
    doi = "10.18653/v1/2022.acl-long.556",
    pages = "8086--8098"
}

@inproceedings{nie-etal-2020-adversarial,
    title = "Adversarial {NLI}: A New Benchmark for Natural Language Understanding",
    author = "Nie, Yixin  and Williams, Adina  and Dinan, Emily  and Bansal, Mohit  and Weston, Jason  and Kiela, Douwe",
    editor = "Jurafsky, Dan  and Chai, Joyce  and Schluter, Natalie  and Tetreault, Joel",
    booktitle = "Proceedings of the 58th Annual Meeting of the Association for Computational Linguistics",
    month = jul,
    year = "2020",
    address = "Online",
    publisher = "Association for Computational Linguistics",
    url = "https://aclanthology.org/2020.acl-main.441/",
    doi = "10.18653/v1/2020.acl-main.441",
    pages = "4885--4901"
}

@misc{saito2023verbosity,
  title         = {Verbosity Bias in Preference Labeling by Large Language Models},
  author        = {Saito, Keita and Wachi, Akifumi and Wataoka, Koki and Akimoto, Youhei},
  year          = {2023},
  eprint        = {2310.10076},
  archivePrefix = {arXiv},
  url           = {https://arxiv.org/abs/2310.10076}
}

@inproceedings{panickssery2024llm,
  title     = {{LLM} Evaluators Recognize and Favor Their Own Generations},
  author    = {Panickssery, Arjun and Bowman, Samuel R. and Feng, Shi},
  booktitle = {Advances in Neural Information Processing Systems},
  volume    = {37},
  pages     = {68772--68802},
  year      = {2024},
  url       = {https://proceedings.neurips.cc/paper_files/paper/2024/hash/7f1f0218e45f5414c79c0679633e47bc-Abstract-Conference.html}
}

@misc{stureborg2024inconsistent,
  title         = {Large Language Models are Inconsistent and Biased Evaluators},
  author        = {Stureborg, Rickard and Alikaniotis, Dimitris and Suhara, Yoshi},
  year          = {2024},
  eprint        = {2405.01724},
  archivePrefix = {arXiv},
  url           = {https://arxiv.org/abs/2405.01724}
}

@misc{li2025scoring,
  title         = {Evaluating Scoring Bias in {LLM-as-a-Judge}},
  author        = {Li, Qingquan and Dou, Shaoyu and Shao, Kailai and Chen, Chao and Hu, Haixiang},
  year          = {2025},
  eprint        = {2506.22316},
  archivePrefix = {arXiv},
  url           = {https://arxiv.org/abs/2506.22316}
}

@inproceedings{zhang2015character,
  title     = {Character-level Convolutional Networks for Text Classification},
  author    = {Zhang, Xiang and Zhao, Junbo and LeCun, Yann},
  booktitle = {Advances in Neural Information Processing Systems},
  volume    = {28},
  year      = {2015},
  url       = {https://proceedings.neurips.cc/paper_files/paper/2015/hash/250cf8b51c773f3f8dc8b4be867a9a02-Abstract.html}
}

@inproceedings{cer-etal-2017-semeval,
    title = "{S}em{E}val-2017 Task 1: Semantic Textual Similarity Multilingual and Crosslingual Focused Evaluation",
    author = "Cer, Daniel  and Diab, Mona  and Agirre, Eneko  and Lopez-Gazpio, I{\~n}igo  and Specia, Lucia",
    editor = "Bethard, Steven  and Carpuat, Marine  and Apidianaki, Marianna  and Mohammad, Saif M.  and Cer, Daniel  and Jurgens, David",
    booktitle = "Proceedings of the 11th International Workshop on Semantic Evaluation ({S}em{E}val-2017)",
    month = aug,
    year = "2017",
    address = "Vancouver, Canada",
    publisher = "Association for Computational Linguistics",
    url = "https://aclanthology.org/S17-2001/",
    doi = "10.18653/v1/S17-2001",
    pages = "1--14"
}

@inproceedings{borkan2019nuanced,
  title     = {Nuanced Metrics for Measuring Unintended Bias with Real Data for Text Classification},
  author    = {Borkan, Daniel and Dixon, Lucas and Sorensen, Jeffrey and Thain, Nithum and Vasserman, Lucy},
  booktitle = {Companion Proceedings of The 2019 World Wide Web Conference},
  series    = {WWW '19},
  pages     = {491--500},
  year      = {2019},
  publisher = {ACM},
  doi       = {10.1145/3308560.3317593}
}

@inproceedings{gehman-etal-2020-realtoxicityprompts,
    title = "{R}eal{T}oxicity{P}rompts: Evaluating Neural Toxic Degeneration in Language Models",
    author = "Gehman, Samuel  and Gururangan, Suchin  and Sap, Maarten  and Choi, Yejin  and Smith, Noah A.",
    editor = "Cohn, Trevor  and He, Yulan  and Liu, Yang",
    booktitle = "Findings of the Association for Computational Linguistics: EMNLP 2020",
    month = nov,
    year = "2020",
    address = "Online",
    publisher = "Association for Computational Linguistics",
    url = "https://aclanthology.org/2020.findings-emnlp.301/",
    doi = "10.18653/v1/2020.findings-emnlp.301",
    pages = "3356--3369"
}

@inproceedings{chang2026balora,
  title     = {{BA-LoRA}: Bias-Alleviating Low-Rank Adaptation to Mitigate Catastrophic Inheritance in Large Language Models},
  author    = {Chang, Yupeng and Chang, Yi and Wu, Yuan},
  booktitle = {The Fourteenth International Conference on Learning Representations},
  year      = {2026},
  url       = {https://openreview.net/forum?id=q0X9SiXiRO}
}

@inproceedings{hu2022lora,
  title     = {{LoRA}: Low-Rank Adaptation of Large Language Models},
  author    = {Hu, Edward J. and Shen, Yelong and Wallis, Phillip and Allen-Zhu, Zeyuan and Li, Yuanzhi and Wang, Shean and Wang, Lu and Chen, Weizhu},
  booktitle = {International Conference on Learning Representations},
  year      = {2022},
  url       = {https://openreview.net/forum?id=nZeVKeeFYf9}
}

@misc{palmer2026kev,
  title        = {Kev},
  author       = {Palmer, Jared},
  year         = {2026},
  howpublished = {\url{https://github.com/jaredpalmer/kev}},
  note         = {GitHub repository. Accessed 2026-09-28}
}

@misc{typesafe2026api,
  title        = {{API} Reference: {TypeSafe System One API}},
  author       = {{TypeSafe AI}},
  year         = {2026},
  howpublished = {\url{https://docs.typesafe.ai/api}},
  note         = {Accessed 2026-09-28}
}

@misc{hume2026jev,
  title        = {Jev's Architecture Unmasked},
  author       = {Hume, Archer},
  year         = {2026},
  howpublished = {\url{https://archerhume.com/posts/jevs-architecture-unmasked/}},
  note         = {Blog post, 17 September 2026}
}

@misc{rao2026jevrubric,
  title         = {{JEV} vs. {LLMs} as Rubric Judges: Cheaper, Faster, and Wrong in the Same Places},
  author        = {Rao, Delip and Callison-Burch, Chris},
  year          = {2026},
  eprint        = {2609.29769},
  archivePrefix = {arXiv},
  url           = {https://arxiv.org/abs/2609.29769}
}

@misc{li2026jevjudge,
  title         = {{JEV}-as-a-Judge: Accept When Confident, Escalate When Unsure},
  author        = {Li, Yubo and Miao, Yidi and Krishnan, Ramayya and Padman, Rema},
  year          = {2026},
  eprint        = {2609.26550},
  archivePrefix = {arXiv},
  url           = {https://arxiv.org/abs/2609.26550}
}

@misc{qwen3.5,
  title  = {{Qwen3.5}: Towards Native Multimodal Agents},
  author = {{Qwen Team}},
  year   = {2026},
  howpublished = {\url{https://qwen.ai/blog?id=qwen3.5}},
  note   = {Blog post, February 2026}
}

@inproceedings{mcauley-leskovec-2013-amateurs,
  title     = {From Amateurs to Connoisseurs: Modeling the Evolution of User Expertise through Online Reviews},
  author    = {McAuley, Julian and Leskovec, Jure},
  booktitle = {Proceedings of the 22nd International Conference on World Wide Web},
  pages     = {897--908},
  year      = {2013},
  doi       = {10.1145/2488388.2488466},
  url       = {https://doi.org/10.1145/2488388.2488466}
}

@inproceedings{potts-etal-2021-dynasent,
  title     = {{DynaSent}: A Dynamic Benchmark for Sentiment Analysis},
  author    = {Potts, Christopher and Wu, Zhengxuan and Geiger, Atticus and Kiela, Douwe},
  booktitle = {Proceedings of the 59th Annual Meeting of the Association for Computational Linguistics and the 11th International Joint Conference on Natural Language Processing (Volume 1: Long Papers)},
  pages     = {2388--2404},
  year      = {2021},
  publisher = {Association for Computational Linguistics},
  doi       = {10.18653/v1/2021.acl-long.186},
  url       = {https://aclanthology.org/2021.acl-long.186/}
}

@inproceedings{poria-etal-2019-meld,
  title     = {{MELD}: A Multimodal Multi-Party Dataset for Emotion Recognition in Conversations},
  author    = {Poria, Soujanya and Hazarika, Devamanyu and Majumder, Navonil and Naik, Gautam and Cambria, Erik and Mihalcea, Rada},
  booktitle = {Proceedings of the 57th Annual Meeting of the Association for Computational Linguistics},
  pages     = {527--536},
  year      = {2019},
  publisher = {Association for Computational Linguistics},
  doi       = {10.18653/v1/P19-1050},
  url       = {https://aclanthology.org/P19-1050/}
}

@article{crossley-etal-2024-persuade,
  title   = {A Large-Scale Corpus for Assessing Written Argumentation: {PERSUADE} 2.0},
  author  = {Crossley, Scott A. and Tian, Yu and Baffour, Perpetual and Franklin, Alex and Benner, Meg and Boser, Ulrich},
  journal = {Assessing Writing},
  volume  = {61},
  pages   = {100865},
  year    = {2024},
  doi     = {10.1016/j.asw.2024.100865},
  url     = {https://doi.org/10.1016/j.asw.2024.100865}
}

@inproceedings{barbieri-etal-2020-tweeteval,
  title     = {{TweetEval}: Unified Benchmark and Comparative Evaluation for Tweet Classification},
  author    = {Barbieri, Francesco and Camacho-Collados, Jose and Espinosa Anke, Luis and Neves, Leonardo},
  booktitle = {Findings of the Association for Computational Linguistics: EMNLP 2020},
  pages     = {1644--1650},
  year      = {2020},
  publisher = {Association for Computational Linguistics},
  doi       = {10.18653/v1/2020.findings-emnlp.148},
  url       = {https://aclanthology.org/2020.findings-emnlp.148/}
}

@inproceedings{mohammad-etal-2018-semeval,
  title     = {{SemEval}-2018 Task 1: Affect in Tweets},
  author    = {Mohammad, Saif and Bravo-Marquez, Felipe and Salameh, Mohammad and Kiritchenko, Svetlana},
  booktitle = {Proceedings of the 12th International Workshop on Semantic Evaluation},
  pages     = {1--17},
  year      = {2018},
  publisher = {Association for Computational Linguistics},
  doi       = {10.18653/v1/S18-1001},
  url       = {https://aclanthology.org/S18-1001/}
}

@misc{hou-etal-2024-amazon,
  title         = {Bridging Language and Items for Retrieval and Recommendation},
  author        = {Hou, Yupeng and Li, Jiacheng and He, Zhankui and Yan, An and Chen, Xiusi and McAuley, Julian},
  year          = {2024},
  eprint        = {2403.03952v1},
  archivePrefix = {arXiv},
  url           = {https://arxiv.org/abs/2403.03952v1}
}

@misc{grano-etal-2017-appreviews,
  title        = {Software Applications User Reviews},
  author       = {Grano, Giovanni and Di Sorbo, Andrea and Mercaldo, Francesco and Visaggio, Corrado A. and Canfora, Gerardo and Panichella, Sebastiano},
  year         = {2017},
  howpublished = {Dataset},
  url          = {https://huggingface.co/datasets/sealuzh/app_reviews}
}

@inproceedings{bagher-zadeh-etal-2018-multimodal,
  title     = {Multimodal Language Analysis in the Wild: {CMU-MOSEI} Dataset and Interpretable Dynamic Fusion Graph},
  author    = {Bagher Zadeh, AmirAli and Liang, Paul Pu and Poria, Soujanya and Cambria, Erik and Morency, Louis-Philippe},
  booktitle = {Proceedings of the 56th Annual Meeting of the Association for Computational Linguistics (Volume 1: Long Papers)},
  pages     = {2236--2246},
  year      = {2018},
  publisher = {Association for Computational Linguistics},
  doi       = {10.18653/v1/P18-1208},
  url       = {https://aclanthology.org/P18-1208/}
}

@inproceedings{wan-mcauley-2018-goodreads,
  title     = {Item Recommendation on Monotonic Behavior Chains},
  author    = {Wan, Mengting and McAuley, Julian},
  booktitle = {Proceedings of the 12th ACM Conference on Recommender Systems},
  pages     = {86--94},
  year      = {2018},
  doi       = {10.1145/3240323.3240369},
  url       = {https://doi.org/10.1145/3240323.3240369}
}

@inproceedings{wang-etal-2024-helpsteer,
  title     = {{HelpSteer}: Multi-Attribute Helpfulness Dataset for {SteerLM}},
  author    = {Wang, Zhilin and Dong, Yi and Zeng, Jiaqi and Adams, Virginia and Sreedhar, Makesh Narsimhan and Egert, Daniel and Delalleau, Olivier and Scowcroft, Jane and Kant, Neel and Swope, Aidan and Kuchaiev, Oleksii},
  booktitle = {Proceedings of the 2024 Conference of the North American Chapter of the Association for Computational Linguistics: Human Language Technologies (Volume 1: Long Papers)},
  pages     = {3371--3384},
  year      = {2024},
  publisher = {Association for Computational Linguistics},
  doi       = {10.18653/v1/2024.naacl-long.185},
  url       = {https://aclanthology.org/2024.naacl-long.185/}
}

@inproceedings{wang-etal-2024-helpsteer2,
  title     = {{HelpSteer2}: Open-Source Dataset for Training Top-Performing Reward Models},
  author    = {Wang, Zhilin and Dong, Yi and Delalleau, Olivier and Zeng, Jiaqi and Shen, Gerald and Egert, Daniel and Zhang, Jimmy J. and Sreedhar, Makesh Narsimhan and Kuchaiev, Oleksii},
  booktitle = {Advances in Neural Information Processing Systems},
  volume    = {37},
  year      = {2024},
  url       = {https://proceedings.neurips.cc/paper_files/paper/2024/hash/02fd91a387a6a5a5751e81b58a75af90-Abstract-Datasets_and_Benchmarks_Track.html}
}

@inproceedings{gretz-etal-2020-argument,
  title     = {A Large-Scale Dataset for Argument Quality Ranking: Construction and Analysis},
  author    = {Gretz, Shai and Friedman, Roni and Cohen-Karlik, Edo and Toledo, Assaf and Lahav, Dan and Aharonov, Ranit and Slonim, Noam},
  booktitle = {Proceedings of the AAAI Conference on Artificial Intelligence},
  volume    = {34},
  pages     = {7805--7813},
  year      = {2020},
  doi       = {10.1609/aaai.v34i05.6285},
  url       = {https://doi.org/10.1609/aaai.v34i05.6285}
}

@inproceedings{aly-atiya-2013-labr,
  title     = {{LABR}: A Large Scale Arabic Book Reviews Dataset},
  author    = {Aly, Mohamed and Atiya, Amir},
  booktitle = {Proceedings of the 51st Annual Meeting of the Association for Computational Linguistics (Volume 2: Short Papers)},
  pages     = {494--498},
  year      = {2013},
  publisher = {Association for Computational Linguistics},
  url       = {https://aclanthology.org/P13-2088/}
}

@inproceedings{socher-etal-2013-sst,
  title     = {Recursive Deep Models for Semantic Compositionality Over a Sentiment Treebank},
  author    = {Socher, Richard and Perelygin, Alex and Wu, Jean and Chuang, Jason and Manning, Christopher D. and Ng, Andrew and Potts, Christopher},
  booktitle = {Proceedings of the 2013 Conference on Empirical Methods in Natural Language Processing},
  pages     = {1631--1642},
  year      = {2013},
  publisher = {Association for Computational Linguistics},
  url       = {https://aclanthology.org/D13-1170/}
}

@misc{argilla-2022-tripadvisor,
  title        = {TripAdvisor Hotel Reviews},
  author       = {{Argilla}},
  year         = {2022},
  howpublished = {Hugging Face dataset},
  url          = {https://huggingface.co/datasets/argilla/tripadvisor-hotel-reviews},
  note         = {Accessed 2026-09-29}
}

@article{alam2016tripadvisor,
  title   = {Joint Multi-Grain Topic Sentiment: Modeling Semantic Aspects for Online Reviews},
  author  = {Alam, Hijbul and Ryu, Woo-Jong and Lee, SangKeun},
  journal = {Information Sciences},
  volume  = {339},
  pages   = {206--223},
  year    = {2016},
  doi     = {10.1016/j.ins.2016.01.013}
}

@inproceedings{cui-etal-2024-ultrafeedback,
  title     = {{UltraFeedback}: Boosting Language Models with Scaled AI Feedback},
  author    = {Cui, Ganqu and Yuan, Lifan and Ding, Ning and Yao, Guanming and He, Bingxiang and Zhu, Wei and Ni, Yuan and Xie, Guotong and Xie, Ruobing and Lin, Yankai and Liu, Zhiyuan and Sun, Maosong},
  booktitle = {Proceedings of the 41st International Conference on Machine Learning},
  volume    = {235},
  pages     = {9722--9744},
  year      = {2024},
  url       = {https://proceedings.mlr.press/v235/cui24f.html}
}

@article{crossley-etal-2025-asap,
  title   = {A Large-Scale Corpus for Assessing Source-Based Writing Quality: {ASAP} 2.0},
  author  = {Crossley, Scott A. and Baffour, Perpetual and Burleigh, L. and King, Jules},
  journal = {Assessing Writing},
  volume  = {65},
  pages   = {100954},
  year    = {2025},
  doi     = {10.1016/j.asw.2025.100954},
  url     = {https://doi.org/10.1016/j.asw.2025.100954}
}

@misc{smallari-2024-openreview,
  title        = {OpenReview {ICLR} Peer Reviews},
  author       = {{smallari}},
  year         = {2025},
  howpublished = {Hugging Face dataset},
  url          = {https://huggingface.co/datasets/smallari/openreview-iclr-peer-reviews},
  note         = {Accessed 2026-09-29}
}

@inproceedings{hossain-etal-2019-humicroedit,
  title     = {``President Vows to Cut \textless Taxes\textgreater{} Hair'': Dataset and Analysis of Creative Text Editing for Humorous Headlines},
  author    = {Hossain, Nabil and Krumm, John and Gamon, Michael},
  booktitle = {Proceedings of the 2019 Conference of the North American Chapter of the Association for Computational Linguistics: Human Language Technologies, Volume 1 (Long and Short Papers)},
  pages     = {133--142},
  year      = {2019},
  publisher = {Association for Computational Linguistics},
  doi       = {10.18653/v1/N19-1012},
  url       = {https://aclanthology.org/N19-1012/}
}

@inproceedings{maas-etal-2011-imdb,
  title     = {Learning Word Vectors for Sentiment Analysis},
  author    = {Maas, Andrew L. and Daly, Raymond E. and Pham, Peter T. and Huang, Dan and Ng, Andrew Y. and Potts, Christopher},
  booktitle = {Proceedings of the 49th Annual Meeting of the Association for Computational Linguistics: Human Language Technologies},
  pages     = {142--150},
  year      = {2011},
  publisher = {Association for Computational Linguistics},
  url       = {https://aclanthology.org/P11-1015/}
}

@inproceedings{sachdeva-etal-2022-hate,
  title     = {The Measuring Hate Speech Corpus: Leveraging Rasch Measurement Theory for Data Perspectivism},
  author    = {Sachdeva, Pratik S. and Barreto, Renata and Bacon, Geoff and Sahn, Alexander and von Vacano, Claudia and Kennedy, Chris},
  editor    = {Abercrombie, Gavin and Basile, Valerio and Tonelli, Sara and Rieser, Verena and Uma, Alexandra},
  booktitle = {Proceedings of the 1st Workshop on Perspectivist Approaches to {NLP} @{LREC}2022},
  month     = jun,
  pages     = {83--94},
  year      = {2022},
  address   = {Marseille, France},
  publisher = {European Language Resources Association},
  url       = {https://aclanthology.org/2022.nlperspectives-1.11/}
}

@inproceedings{danescu-niculescu-mizil-etal-2013-politeness,
  title     = {A Computational Approach to Politeness with Application to Social Factors},
  author    = {Danescu-Niculescu-Mizil, Cristian and Sudhof, Moritz and Jurafsky, Dan and Leskovec, Jure and Potts, Christopher},
  booktitle = {Proceedings of the 51st Annual Meeting of the Association for Computational Linguistics (Volume 1: Long Papers)},
  pages     = {250--259},
  year      = {2013},
  publisher = {Association for Computational Linguistics},
  url       = {https://aclanthology.org/P13-1025/}
}

@article{brysbaert-etal-2014-concreteness,
  title   = {Concreteness Ratings for 40 Thousand Generally Known English Word Lemmas},
  author  = {Brysbaert, Marc and Warriner, Amy Beth and Kuperman, Victor},
  journal = {Behavior Research Methods},
  volume  = {46},
  number  = {3},
  pages   = {904--911},
  year    = {2014},
  doi     = {10.3758/s13428-013-0403-5},
  url     = {https://doi.org/10.3758/s13428-013-0403-5}
}

@inproceedings{buechel-hahn-2017-emobank,
  title     = {{EmoBank}: Studying the Impact of Annotation Perspective and Representation Format on Dimensional Emotion Analysis},
  author    = {Buechel, Sven and Hahn, Udo},
  booktitle = {Proceedings of the 15th Conference of the European Chapter of the Association for Computational Linguistics: Volume 2, Short Papers},
  pages     = {578--585},
  year      = {2017},
  publisher = {Association for Computational Linguistics},
  url       = {https://aclanthology.org/E17-2092/}
}

@misc{kebez-2024-pitchfork,
  title        = {Pitchfork Album Reviews},
  author       = {{Kebez}},
  year         = {2025},
  howpublished = {Hugging Face dataset},
  url          = {https://huggingface.co/datasets/Kebez/Pitchfork-Album-Reviews},
  note         = {Accessed 2026-09-29}
}

@inproceedings{ethayarajh-etal-2022-shp,
  title     = {Understanding Dataset Difficulty with \emph{V}-Usable Information},
  author    = {Ethayarajh, Kawin and Choi, Yejin and Swayamdipta, Swabha},
  booktitle = {Proceedings of the 39th International Conference on Machine Learning},
  volume    = {162},
  pages     = {5988--6008},
  year      = {2022},
  url       = {https://proceedings.mlr.press/v162/ethayarajh22a.html}
}

@inproceedings{graesser-etal-2018-drugs,
  title     = {Aspect-Based Sentiment Analysis of Drug Reviews Applying Cross-Domain and Cross-Data Learning},
  author    = {Gr{\"a}{\ss}er, Felix and Kallumadi, Surya and Malberg, Hagen and Zaunseder, Sebastian},
  booktitle = {Proceedings of the 2018 International Conference on Digital Health},
  pages     = {121--125},
  year      = {2018},
  doi       = {10.1145/3194658.3194677},
  url       = {https://doi.org/10.1145/3194658.3194677}
}

@misc{wada-2025-metacritic,
  title        = {Metacritic Games Reviews Dataset},
  author       = {{Wada1}},
  year         = {2025},
  howpublished = {Hugging Face dataset},
  url          = {https://huggingface.co/datasets/Wada1/Metacritic_Games_Reviews_Dataset},
  note         = {Accessed 2026-09-29}
}

@misc{spawn99-2025-wine,
  title        = {Wine Reviews},
  author       = {{spawn99}},
  year         = {2025},
  howpublished = {Hugging Face dataset},
  url          = {https://huggingface.co/datasets/spawn99/wine-reviews},
  note         = {Accessed 2026-09-29}
}

@misc{lightblue-2024-text-ratings,
  title        = {Text Ratings},
  author       = {{Lightblue}},
  year         = {2024},
  howpublished = {Hugging Face dataset},
  url          = {https://huggingface.co/datasets/lightblue/text_ratings},
  note         = {Accessed 2026-09-29}
}

@inproceedings{specia-etal-2020-wmt,
  title     = {Findings of the {WMT} 2020 Shared Task on Quality Estimation},
  author    = {Specia, Lucia and Blain, Fr{\'e}d{\'e}ric and Fomicheva, Marina and Fonseca, Erick and Chaudhary, Vishrav and Guzm{\'a}n, Francisco and Martins, Andr{\'e} F. T.},
  booktitle = {Proceedings of the Fifth Conference on Machine Translation},
  pages     = {743--764},
  year      = {2020},
  publisher = {Association for Computational Linguistics},
  doi       = {10.18653/v1/2020.wmt-1.79},
  url       = {https://aclanthology.org/2020.wmt-1.79/}
}

@misc{lambert-etal-2023-stackexchange,
  title        = {Hugging Face {H4} Stack Exchange Preference Dataset},
  author       = {Lambert, Nathan and Tunstall, Lewis and Rajani, Nazneen and Thrush, Tristan},
  year         = {2023},
  howpublished = {Hugging Face dataset},
  url          = {https://huggingface.co/datasets/HuggingFaceH4/stack-exchange-preferences}
}

@misc{coucke-etal-2018-snips,
  title         = {Snips Voice Platform: An Embedded Spoken Language Understanding System for Private-by-Design Voice Interfaces},
  author        = {Coucke, Alice and Saade, Alaa and Ball, Adrien and Bluche, Th{\'e}odore and Caulier, Alexandre and Leroy, David and Doumouro, Cl{\'e}ment and Gisselbrecht, Thibault and Caltagirone, Francesco and Lavril, Thibaut and Primet, Ma{\"e}l and Dureau, Joseph},
  year          = {2018},
  eprint        = {1805.10190},
  archivePrefix = {arXiv},
  url           = {https://arxiv.org/abs/1805.10190}
}

@inproceedings{chalkidis-etal-2022-lexglue,
  title     = {{LexGLUE}: A Benchmark Dataset for Legal Language Understanding in English},
  author    = {Chalkidis, Ilias and Jana, Abhik and Hartung, Dirk and Bommarito, Michael and Androutsopoulos, Ion and Katz, Daniel and Aletras, Nikolaos},
  booktitle = {Proceedings of the 60th Annual Meeting of the Association for Computational Linguistics (Volume 1: Long Papers)},
  pages     = {4310--4330},
  year      = {2022},
  publisher = {Association for Computational Linguistics},
  doi       = {10.18653/v1/2022.acl-long.297},
  url       = {https://aclanthology.org/2022.acl-long.297/}
}
\appendix
\clearpage
\section*{Appendix}

The appendices provide further interpretation and details of the experimental protocol.
Appendix~\ref{sec:discussion} synthesizes the findings and their implications for evaluating direct-decision models.
Appendix~\ref{sec:appendix-datasets} lists the nine models, experiment coverage, 40 datasets, candidate-order protocol, and artifact scope.
Appendix~\ref{sec:appendix-metrics} defines the metrics and aggregation rules.
Appendix~\ref{sec:appendix-construction} specifies both controlled constructions: scale refinement (\S\ref{sec:kscaling}) and fixed-count ordinal spacing (\S\ref{sec:distance}).
Appendix~\ref{sec:appendix-posttraining} provides the source-selection rule and complete post-training configuration (\S\ref{sec:posttraining}).
Appendix~\ref{sec:appendix-results} reports main-panel variation, candidate-order diagnostics, scale-refinement endpoints, complete spacing results, and post-training analyses.

\section{Extended Discussion}
\label{sec:discussion}

\paragraph{Correctness, scale utilization, and order sensitivity are distinct properties.}
The main panel and position intervention show why candidate-space compression cannot be reduced to one favored label or presentation position.
Preferred anchors vary across models and tasks, while many individual predictions change under randomization without a comparable change in accuracy.
These findings distinguish aggregate scale use from both item-level correctness and order sensitivity (\S\ref{sec:broad} and \S\ref{sec:position}).
Evaluating any one property alone can therefore miss a different form of decision bias.

\paragraph{Probability coverage does not guarantee decision-level utilization.}
Under scale refinement, most models retain broad average probability coverage while final decisions increasingly concentrate and top-1 margins narrow (\S\ref{sec:kscaling}).
This pattern is consistent with repeated ranking advantages for particular levels, rather than complete exclusion of other candidates from probability space.
JEV also exhibits declining soft utilization, indicating that probability concentration and restricted argmax decisions can coexist.

\paragraph{Ordinal structure matters beyond local candidate spacing.}
Nominal tasks and ordinal scale refinement have different utilization profiles, but this contrast does not isolate semantic coupling from task difficulty.
The fixed-count intervention further shows that wider ordinal gaps can improve correctness without restoring utilization (\S\ref{sec:distance}).
Changing spacing also changes the available distractors, so rank distance is not a general measure of semantic coupling.
Together, the interventions implicate how models use ordinal alternatives, not a universal rule that closer candidates necessarily produce greater compression.

\paragraph{Post-training makes scale utilization an actionable target.}
BA-LoRA improves utilization, distributional alignment, and correctness on held-out items from the eight targeted sources (\S\ref{sec:posttraining}).
Transfer differs by model size: aggregate utilization improves on unseen ordinal sources at 4B but worsens at 0.8B.
These results establish that KEV's scale use is adaptable, without implying uniform transfer or identifying the cause of JEV's behavior.

\paragraph{Evaluation should audit both decisions and distributions.}
For rubric judges and first-stage evaluators in confidence-based cascades \citep{li2026jevjudge}, accuracy should be accompanied by $D_R$, TVD, and candidate-order sensitivity.
Same-item scale refinement and fixed-count spacing then test complementary aspects of candidate use.
Low $D_R$ indicates matched effective support, not necessarily correct decisions; joint reporting distinguishes useful scale recovery from indiscriminate distribution spreading.

\section{Datasets and Data Collection}
\label{sec:appendix-datasets}

\subsection{Models and Experiment Coverage}
\label{sec:appendix-models}

We use one hosted JEV version, three released KEV checkpoints, and five StartLux Decision checkpoints.
Table~\ref{tab:experiment-coverage} distinguishes unique source items from constructed queries and records the coverage of each experiment.
All nine models complete the main, position, and candidate-count protocols.
For the seven-point spacing experiment, KEV and StartLux cover four sources, whereas JEV covers the three-source panel excluding Word Concreteness.
Comparisons across all nine models therefore use only these three shared sources.

\begin{table*}[t]
  \centering
  \small
  \setlength{\tabcolsep}{5pt}
  \begin{tabular}{p{0.18\textwidth}p{0.20\textwidth}rrp{0.30\textwidth}}
    \toprule
    \textbf{Experiment} & \textbf{Model coverage} & \textbf{Sources} & \textbf{Queries/model} & \textbf{Source items and reuse} \\
    \midrule
    Main evaluation & All nine models & 40 & 200,000 & 5,000 items per source \\
    Candidate order & All nine models & 9 & 90,000 & 45,000 main-panel items, two new orders each \\
    Candidate count & All nine models & 4 & 127,400 & 1,400 items per source; $K=2,\ldots,14$; two ordinal binnings \\
    Ordinal spacing & KEV (3); StartLux (5) & 4 & 138,894 & 19,842 items, seven fixed-$K$ candidate sets each \\
    Ordinal spacing & JEV 1.13 & 3 & 104,118 & 14,874 of the same items; excludes Word Concreteness \\
    Post-training evaluation & KEV 0.8/4B, before and after BA-LoRA & 40 & 200,000 & Frozen main panel; 32,000 disjoint training items \\
    \bottomrule
  \end{tabular}
  \caption{\label{tab:experiment-coverage}
    \textbf{Experiment coverage.} Queries are constructed inputs, not unique texts or billed requests. Main-panel successes: 199,999 per original KEV checkpoint; 200,000 for JEV/StartLux. Original-order decisions and identical spacing endpoints are reusable.
  }
\end{table*}

\paragraph{Model interfaces.}
The hosted endpoint identifies JEV as \texttt{typesafe/jev-1.13-20260917}.
The KEV checkpoints are \texttt{jaredpalmer/kev-0.8b}, \texttt{kev-4b}, and \texttt{kev-9b}; each combines its Qwen3.5 backbone with the released decision adapter and pointer head.
StartLux Decision is evaluated at 0.8B, 2B, 4B, 9B, and 27B using its native candidate-scoring interface.
KEV and StartLux return candidate probabilities directly rather than generating a textual verdict.
The model repositories are linked in the main-text model description.

\paragraph{Recorded inference resources.}
The query counts in Table~\ref{tab:experiment-coverage} describe constructed evaluations, not provider-billed calls or GPU-hours. JEV receives 521,518 constructed queries across the four study panels plus 6,400 ANLI items; the hosted service does not expose its internal GPU usage. Saved StartLux final-run summaries sum to 21.90 hours of elapsed time across five sizes and four panels. Elapsed time for all KEV baseline inference and exploratory runs was not consistently logged.
Local KEV and StartLux inference used NVIDIA GeForce RTX 4080 SUPER GPUs, except for StartLux-27B, which used an NVIDIA RTX PRO 6000 GPU; the hosted JEV hardware was not disclosed.

\subsection{Main Evaluation Panel}

Table~\ref{tab:datasets} groups the 40 datasets in the main evaluation panel by task type and number of candidates $K$.

\begin{table}[!ht]
  \centering
  \small
  \begin{tabular}{lp{0.80\linewidth}}
    \toprule
    $K$ & \textbf{Datasets} \\
    \midrule
    \multicolumn{2}{l}{\emph{Ordinal}} \\
    3 & BeerAdvocate Reviews, DynaSent, MELD Sentiment, PERSUADE 2.0 Argument Effectiveness, TweetEval Sentiment \\
    4 & SemEval-2018 Emotion Intensity \\
    5 & Amazon Reviews 2023, App Reviews, CMU-MOSEI, Goodreads, HelpSteer, HelpSteer2 Correctness, IBM Argument Quality, LABR, SST-5, TripAdvisor, UltraFeedback \\
    6 & ASAP 2.0, OpenReview Recommendation \\
    7 & Civil Comments Toxicity \\
    8 & Humicroedit Funniness, IMDb Ratings, Measuring Hate Speech \\
    9 & Stanford Politeness, Word Concreteness \\
    10 & EmoBank Valence, Pitchfork Albums, SHP Community Approval, Drugs.com \\
    11 & Metacritic Games, STS Benchmark Similarity, Wine Quality \\
    12 & WebText Quality, WMT20 Translation Quality \\
    14 & RealToxicity Continuation, StackExchange Answer Score \\
    \midrule
    \multicolumn{2}{l}{\emph{Nominal}} \\
    7--14 & SNIPS (7), Yahoo Answers (10), SCOTUS (13), DBpedia (14) \\
    \bottomrule
  \end{tabular}
  \caption{\label{tab:datasets}
    \textbf{Main-panel datasets} by task type and candidate count $K$; 5,000 items per dataset.
  }
\end{table}

\paragraph{Dataset sources.}
The panel draws on BeerAdvocate Reviews~\citep{mcauley-leskovec-2013-amateurs},
DynaSent~\citep{potts-etal-2021-dynasent}, MELD~\citep{poria-etal-2019-meld},
PERSUADE~2.0~\citep{crossley-etal-2024-persuade}, TweetEval~\citep{barbieri-etal-2020-tweeteval},
SemEval-2018 Task~1~\citep{mohammad-etal-2018-semeval}, Amazon Reviews'23~\citep{hou-etal-2024-amazon},
App Reviews~\citep{grano-etal-2017-appreviews}, CMU-MOSEI~\citep{bagher-zadeh-etal-2018-multimodal},
Goodreads Reviews~\citep{wan-mcauley-2018-goodreads}, HelpSteer~\citep{wang-etal-2024-helpsteer},
HelpSteer2~\citep{wang-etal-2024-helpsteer2}, IBM Argument Quality~\citep{gretz-etal-2020-argument},
LABR~\citep{aly-atiya-2013-labr}, SST-5~\citep{socher-etal-2013-sst},
TripAdvisor Hotel Reviews~\citep{alam2016tripadvisor,argilla-2022-tripadvisor}, UltraFeedback~\citep{cui-etal-2024-ultrafeedback},
ASAP~2.0~\citep{crossley-etal-2025-asap}, OpenReview ICLR reviews~\citep{smallari-2024-openreview},
Civil Comments~\citep{borkan2019nuanced}, Humicroedit~\citep{hossain-etal-2019-humicroedit},
IMDb Reviews~\citep{maas-etal-2011-imdb}, Measuring Hate Speech~\citep{sachdeva-etal-2022-hate},
Stanford Politeness~\citep{danescu-niculescu-mizil-etal-2013-politeness},
Word Concreteness~\citep{brysbaert-etal-2014-concreteness}, EmoBank~\citep{buechel-hahn-2017-emobank},
Pitchfork Album Reviews~\citep{kebez-2024-pitchfork}, Stanford Human Preferences~\citep{ethayarajh-etal-2022-shp},
Drugs.com Reviews~\citep{graesser-etal-2018-drugs}, Metacritic Games Reviews~\citep{wada-2025-metacritic},
STS Benchmark~\citep{cer-etal-2017-semeval}, Wine Reviews~\citep{spawn99-2025-wine},
WebText Quality Ratings~\citep{lightblue-2024-text-ratings}, WMT20 Quality Estimation~\citep{specia-etal-2020-wmt},
RealToxicityPrompts~\citep{gehman-etal-2020-realtoxicityprompts},
H4 Stack Exchange Preferences~\citep{lambert-etal-2023-stackexchange},
SNIPS~\citep{coucke-etal-2018-snips}, Yahoo Answers and DBpedia~\citep{zhang2015character},
and the SCOTUS subset of LexGLUE~\citep{chalkidis-etal-2022-lexglue}.

\paragraph{Sampling.}
Each dataset contributes 5,000 items, drawn from the train, validation, or test split of its source; splits are pooled where one split is too small.
Sampling is stratified over gold labels and uses seed 20260926 or 20260927, depending on the collection batch.
Within each gold-label stratum, items are drawn uniformly without replacement, and the quota of an under-filled stratum is redistributed to the others.

\paragraph{Language coverage.}
The source material is predominantly English, with Arabic book reviews in LABR and English-to-German source--translation pairs in the sampled WMT20 task. Sampling balances gold labels rather than languages or author demographics. Dataset domains, candidate counts, and source citations are given in Table~\ref{tab:datasets} and the source list above.

\paragraph{Source-specific research-use terms.}
Table~\ref{tab:dataset-licenses} records the license or access statement found for each source. A public copy or a mirror's license tag does not by itself establish rights over underlying third-party text. Accordingly, our review artifact does not redistribute source passages; reconstruction remains subject to each provider's terms.

\begin{table*}[t]
  \centering
  \scriptsize
  \setlength{\tabcolsep}{3pt}
  \renewcommand{\arraystretch}{1.08}
  \caption{\label{tab:dataset-licenses}\textbf{Documented licenses and access terms for all 40 main-panel datasets.} Dataset names link to the source statement or data card used for this audit. ``Mirror'' identifies a repository-level label whose applicability to underlying third-party content is not established; ``not stated'' means we did not verify an explicit dataset-level license. These entries describe available documentation, not a common license granted by this paper.}
  \begin{tabular}{@{}p{0.18\textwidth}p{0.27\textwidth}p{0.18\textwidth}p{0.27\textwidth}@{}}
    \toprule
    \textbf{Dataset} & \textbf{Documented terms} & \textbf{Dataset} & \textbf{Documented terms} \\
    \midrule
    \href{https://www.beeradvocate.com/community/threads/faq-data-requests.681967/}{BeerAdvocate} & Source restricts bulk export & \href{https://github.com/SemEvalWorkshop/humicroedit}{Humicroedit} & License not verified \\
    \href{https://github.com/cgpotts/dynasent/blob/main/dynasent_datasheet.md}{DynaSent} & CC BY 4.0 & \href{https://nlp.stanford.edu/~amaas/data/sentiment/}{IMDb Ratings} & Research release; license not stated \\
    \href{https://github.com/declare-lab/MELD}{MELD} & GPL-3.0 repository; dialogue rights separate & \href{https://huggingface.co/datasets/ucberkeley-dlab/measuring-hate-speech}{Measuring Hate Speech} & CC BY 4.0 \\
    \href{https://github.com/scrosseye/persuade_corpus_2.0/blob/main/README.md}{PERSUADE 2.0} & CC BY-NC-SA 4.0 & \href{https://convokit.cornell.edu/documentation/stack_politeness.html}{Stanford Politeness} & CC BY 4.0 (ConvoKit release) \\
    \href{https://huggingface.co/datasets/cardiffnlp/tweet_eval}{TweetEval} & CC BY 3.0; platform terms also apply & \href{https://huggingface.co/datasets/StephanAkkerman/concreteness-ratings}{Word Concreteness} & MIT (mirror); original terms unverified \\
    \href{https://saifmohammad.com/WebPages/SentimentEmotionLabeledData.html}{SemEval-2018 EI-OC} & Research use with citation; no redistribution & \href{https://github.com/JULIELab/EmoBank}{EmoBank Valence} & CC BY-SA 4.0 \\
    \href{https://huggingface.co/datasets/McAuley-Lab/Amazon-Reviews-2023/discussions/1}{Amazon Reviews 2023} & Research release; license not assigned & \href{https://huggingface.co/datasets/Kebez/Pitchfork-Album-Reviews}{Pitchfork Albums} & CC BY 4.0 (mirror); original rights unverified \\
    \href{https://huggingface.co/datasets/sealuzh/app_reviews}{App Reviews} & License not stated & \href{https://huggingface.co/datasets/stanfordnlp/SHP}{SHP Community Approval} & License not stated; Reddit content \\
    \href{https://github.com/CMU-MultiComp-Lab/CMU-MultimodalSDK/issues/15}{CMU-MOSEI} & Dataset license not stated & \href{https://archive.ics.uci.edu/dataset/462/drug+review+dataset+drugs+com}{Drugs.com} & UCI policy: CC BY 4.0; item terms unverified \\
    \href{https://cseweb.ucsd.edu/~jmcauley/datasets/goodreads.html}{Goodreads} & Academic use only; no redistribution & \href{https://huggingface.co/datasets/Wada1/Metacritic_Games_Reviews_Dataset}{Metacritic Games} & CC BY 4.0 (mirror); original rights unverified \\
    \href{https://huggingface.co/datasets/nvidia/HelpSteer}{HelpSteer} & CC BY 4.0 & \href{https://huggingface.co/datasets/mteb/stsbenchmark-sts}{STS Benchmark} & License not stated on cited mirror \\
    \href{https://huggingface.co/datasets/nvidia/HelpSteer2}{HelpSteer2 Correctness} & CC BY 4.0 & \href{https://huggingface.co/datasets/spawn99/wine-reviews}{Wine Quality} & CC BY-NC-SA 4.0 (mirror claim) \\
    \href{https://huggingface.co/datasets/ibm-research/argument_quality_ranking_30k}{IBM Argument Quality} & CC BY 3.0 & \href{https://huggingface.co/datasets/lightblue/text_ratings}{WebText Quality} & License not stated \\
    \href{https://github.com/mohamedadaly/LABR}{LABR} & GPL-2.0 repository; review rights separate & \href{https://github.com/facebookresearch/mlqe}{WMT20 Translation Quality} & CC BY-SA 4.0 \\
    \href{https://huggingface.co/datasets/SetFit/sst5}{SST-5} & License not stated on cited mirror & \href{https://huggingface.co/datasets/allenai/real-toxicity-prompts}{RealToxicity} & Apache 2.0 card; source-text rights separate \\
    \href{https://huggingface.co/datasets/argilla/tripadvisor-hotel-reviews}{TripAdvisor} & CC BY-NC 4.0 (mirror); review rights unverified & \href{https://stackoverflow.com/help/licensing}{StackExchange Answer Score} & CC BY-SA 2.5/3.0/4.0 by post date \\
    \href{https://huggingface.co/datasets/argilla/ultrafeedback-critique}{UltraFeedback} & \href{https://huggingface.co/datasets/openbmb/UltraFeedback}{MIT upstream}; derivative terms not stated & \href{https://github.com/sonos/nlu-benchmark}{SNIPS} & CC0 1.0 \\
    \href{https://github.com/scrosseye/ASAP_2.0}{ASAP 2.0} & CC BY 4.0 & \href{https://github.com/LC-John/Yahoo-Answers-Topic-Classification-Dataset/blob/master/dataset/readme.txt}{Yahoo Answers} & Noncommercial research; signed agreement required \\
    \href{https://openreview.net/legal/terms}{OpenReview Recommendation} & Public comments: CC BY 4.0 & \href{https://huggingface.co/datasets/coastalcph/lex_glue}{SCOTUS13} & CC BY 4.0 (LexGLUE); mirror terms unstated \\
    \href{https://huggingface.co/datasets/google/civil_comments}{Civil Comments} & CC0 1.0 & \href{https://huggingface.co/datasets/fancyzhx/dbpedia_14}{DBpedia14} & CC BY-SA 3.0 \\
    \bottomrule
  \end{tabular}
\end{table*}

\paragraph{Queries, label mapping, and completed records.}
JEV requests contain no sampling parameters, and candidates are keyed A, B, C, \dots{} in presentation order.
In the second collection batch (10 datasets), inputs whose request exceeds 24,000 bytes are truncated in the middle.
Candidate probabilities are returned per presentation key, and we map them to labels with the recorded permutation of each prompt.
In the position intervention, predictions and probabilities for the original and permuted versions are therefore compared by candidate text, not by position.
Retry records are resolved by sample identifier before scoring, so repeated requests do not count as additional observations.
Responses with a non-\texttt{ok} status are excluded; this affects one of the 200,000 main-experiment items for each KEV model and none for JEV~1.13 or the five StartLux models.
The completed spacing panel contains 104,118 successful unique JEV records and 138,894 for each KEV and StartLux checkpoint.

\paragraph{Artifact scope and reconstruction.}
The data-bearing portion of the anonymous review artifact contains source-indexed sample identifiers, source splits, gold labels, content hashes, construction metadata, and numeric model predictions. It does not redistribute the original dataset passages or upstream model weights; the post-trained KEV checkpoints are likewise absent from this review snapshot. Input reconstruction downloads from the original providers and checks the reconstructed records against stored hashes. Our original analysis code is released under the MIT License. The third-party datasets and model checkpoints retain their respective upstream licenses and access terms; this work does not assign them a common license. Provider locations and recorded source revisions are specified in the artifact configuration. The released instance-level predictions, rather than a fresh download of possibly changed upstream resources, are the fixed records used for the reported analyses.

\paragraph{Use of AI assistants.}
AI assistants supported literature review, programming, and language polishing of author-written manuscript text.

\paragraph{Sensitive source material and intended use.}
Some underlying corpora contain toxic or hateful language, drug reviews, or other user-generated text that may include identifying details. To limit republication of such material, the review artifact excludes the source passages and user-profile fields, retaining only source identifiers, hashes, labels, and numerical outputs needed to verify the analyses. Identifiers can still locate public upstream records and should not be interpreted as anonymizing those records. Rebuilding the inputs requires an explicit download from the source provider under its own terms. Our results describe candidate-selection behavior in benchmark settings; a model that shifts decisions toward a few ordinal levels should not be used without task-specific validation as the sole basis for consequential moderation, educational, health-related, or other high-stakes decisions.

\subsection{Candidate-Order Panel}
\label{sec:appendix-position}

The position intervention uses six ordinal datasets: TripAdvisor Hotel Reviews ($K=5$), HelpSteer2 Correctness ($K=5$), IBM Argument Quality ($K=5$), Civil Comments Toxicity ($K=7$), SHP Community Approval ($K=10$), and RealToxicity Continuation ($K=14$).
The three nominal references are SNIPS ($K=7$), Yahoo Answers ($K=10$), and DBpedia ($K=14$).
This panel spans several scale lengths and includes nominal tasks at matching candidate counts; it is not a random sample of dataset names.

Each dataset reuses its 5,000 main-panel items.
For each item, we retain the text, question, candidate strings, and gold semantic label, and generate two distinct non-identity permutations of the candidate list.
The stored new-to-original permutation maps the displayed gold position, decision, and probability vector back to the canonical labels.
The two permutations produce 90,000 additional queries per model; the 45,000 original-order decisions are reused from the main experiment.
Aggregate randomized metrics are calculated separately for each permutation and then averaged, rather than calculated after pooling the two prediction distributions.
Item-level agreement and probability shift compare the two randomized versions of the same item (Appendix~\ref{sec:appendix-metrics}).

\subsection{Motivating ANLI Observation}
The observation in the introduction (Figure~\ref{fig:anli}) uses JEV~1.13 predictions on all official ANLI development and test items from rounds R1--R3 (6,400 items), with accuracy 74.95\%.
Candidate order is randomized per item and balanced over positions: Neutral is presented first, second, and third for 2,136, 2,132, and 2,132 items, and the gold answer for 2,098, 2,183, and 2,119 items.

\section{Metric Definitions}
\label{sec:appendix-metrics}

Equations~\ref{eq:ueff} and~\ref{eq:soft-utilization} define the utilization metrics used in the main text; this section defines the remaining metrics.
Let $y_i$ and $\hat y_i$ be the gold and predicted labels for item $i=1,\dots,N$, and let $p$ and $q$ be the empirical distributions of predicted and gold labels.
Entropy uses natural logarithms and the convention $0\log 0=0$.

We use $D_R$ consistently in tables, plots, and comparisons, computing the absolute deviation separately for each dataset before macro-averaging.
We report $D_R$ in percentage points, so zero denotes equal predicted and gold effective-support sizes. Its magnitude includes both compression and expansion; compression specifically requires $\exp(H(p))<\exp(H(q))$.
Equal effective support does not imply matching label frequencies or correct predictions; TVD and accuracy assess these separately.

\paragraph{Decision-level metrics.}
Accuracy and distributional divergence are
\begin{equation}
\label{eq:appendix-accuracy-tvd}
\begin{array}{@{}l@{}}
\mathrm{Acc} = \displaystyle\frac{1}{N}\sum_{i=1}^{N}{\bf 1}[\hat y_i=y_i],\\[4pt]
\mathrm{TVD}(p,q) = \displaystyle\frac{1}{2}\sum_{j=1}^{K}|p_j-q_j|.
\end{array}
\end{equation}

\paragraph{Probability metrics.}
For probability metrics, returned nonnegative vectors are divided by their sum and mapped to canonical labels.
Let $\pi_{i,(1)}$ and $\pi_{i,(2)}$ denote the largest and second-largest entries of the resulting vector $\pi_i$.
In addition to $U_{\mathrm{soft}}$ (Equation~\ref{eq:soft-utilization}), we report mean normalized entropy and the mean top-1 margin:
\begin{equation}
\label{eq:appendix-probability-metrics}
\begin{array}{@{}l@{}}
\widetilde H = \displaystyle\frac{1}{N}\sum_{i=1}^{N}\frac{H(\pi_i)}{\log K},\\[4pt]
\mathrm{Margin} = \displaystyle\frac{1}{N}\sum_{i=1}^{N}
  \left(\pi_{i,(1)}-\pi_{i,(2)}\right).
\end{array}
\end{equation}

\paragraph{Ordinal metrics.}
For ordinal labels indexed from $1$ to $K$, we normalize MAE by the maximum possible rank difference, $K-1$, following score-range normalization \citep{mathew2026llmsgrade}.
QWK applies quadratic disagreement weights to Cohen's chance-corrected agreement coefficient \citep{cohen1968weighted,fleisscohen1973equivalence}:
\begin{equation}
\label{eq:appendix-ordinal-metrics}
\begin{array}{@{}l@{}}
\mathrm{nMAE} = \displaystyle\frac{1}{N(K-1)}
  \sum_{i=1}^{N}|\hat y_i-y_i|,\\[6pt]
\mathrm{QWK} = \displaystyle 1-
  \frac{\sum_{a,b=1}^{K}w_{ab}O_{ab}}
       {\sum_{a,b=1}^{K}w_{ab}E_{ab}},\\[6pt]
w_{ab} = \displaystyle\frac{(a-b)^2}{(K-1)^2},
\end{array}
\end{equation}
where $O$ is the observed gold--prediction contingency table and $E$ is its independence-based expectation with the same marginals.

\paragraph{Position sensitivity.}
Let $\hat y_i^{(v)}$ and $\pi_i^{(v)}$ denote the canonical-label prediction and probability vector under randomized order $v\in\{1,2\}$.
The Agreement and Shift columns in Table~\ref{tab:position} are
\begin{equation}
\label{eq:appendix-position-metrics}
\begin{array}{@{}l@{}}
\mathrm{Agreement} = \displaystyle\frac{1}{N}\sum_i {\bf 1}[\hat y_i^{(1)}=\hat y_i^{(2)}],\\[5pt]
\mathrm{Shift} = \displaystyle\frac{1}{2N}\sum_i\sum_{j=1}^{K}
|\pi_{ij}^{(1)}-\pi_{ij}^{(2)}|.
\end{array}
\end{equation}
These measure changes between two orders of the same item, whereas $D_R$ and TVD measure the aggregate label distribution within one order.
For an aggregate metric $M$, the randomized result is $(M^{(1)}+M^{(2)})/2$, computed per dataset before macro-averaging.
In particular, the reported randomized $D_R$ averages the two separately computed absolute deviations; we do not pool predictions across orders or take an absolute value after averaging support ratios.

\paragraph{Label space and aggregation.}
Main-panel and position metrics use the original semantic labels; scale-refinement metrics use the canonical bins at each $K$.
For fixed-count spacing, candidates retain their original level names but differ across items.
We therefore map each displayed set to its relative ascending ranks $1,\ldots,K$ before forming $p$ and $q$.
Presentation position and original absolute level are not the label space of the spacing tables.
The relative gold ranks are exactly balanced, giving $\exp(H(q))=K$ and hence $D_R=100(1-U_{\mathrm{arg}})$ in this experiment.
All multi-dataset summaries are equal-weight means of dataset-level metrics, not metrics of pooled predictions.
Each reported model--protocol result is computed from one frozen prediction set, not an average over repeated model runs.
Accuracy, utilization, TVD, and nMAE are displayed as percentages; $D_R$, Margin, and Shift use percentage points, while QWK and normalized entropy are unscaled.

\section{Controlled Candidate Construction}
\label{sec:appendix-construction}

\subsection{Candidate-Count Scale Refinement}
\label{sec:appendix-k-construction}

The scale-refinement panel uses a frozen subset of 1,400 items per source from the main evaluation sample, with construction seed 20260927.
The same source items are reused at every $K=2,\ldots,14$ and, for the three ordinal sources, under both binning rules.
Thus, the ordinal panel contains $3\times1{,}400\times13\times2=109{,}200$ queries, and the DBpedia reference adds $1{,}400\times13=18{,}200$, totaling 127,400 per model.
The 127,400 queries represent 5,600 source items, not 127,400 independent texts.

\begin{table*}[t]
  \centering
  \small
  \textbf{(a) Representative candidate strings}\par\smallskip
  \begin{tabular}{cp{0.42\textwidth}p{0.42\textwidth}}
    \toprule
    $K$ & \textbf{Equal-width examples} & \textbf{Equal-frequency examples} \\
    \midrule
    2 & Level 01 of 02: $0\le s<0.5$; Level 02 of 02: $0.5\le s\le1$
      & Level 01 of 02: $0\le s<0.503333$; Level 02 of 02: $0.503333\le s\le1$ \\
    5 & Level 01 of 05: $0\le s<0.2$; Level 03 of 05: $0.4\le s<0.6$; Level 05 of 05: $0.8\le s\le1$
      & Level 01 of 05: $0\le s<0.176357$; Level 03 of 05: $0.402703\le s<0.598387$; Level 05 of 05: $0.801316\le s\le1$ \\
    10 & Level 01 of 10: $0\le s<0.1$; Level 05 of 10: $0.4\le s<0.5$; Level 10 of 10: $0.9\le s\le1$
       & Level 01 of 10: $0\le s<0.101282$; Level 05 of 10: $0.402703\le s<0.503333$; Level 10 of 10: $0.893778\le s\le1$ \\
    14 & Level 01 of 14: $0\le s<0.071429$; Level 07 of 14: $0.428571\le s<0.5$; Level 14 of 14: $0.928571\le s\le1$
       & Level 01 of 14: $0\le s<0.101282$; Level 07 of 14: $0.426569\le s<0.503333$; Level 14 of 14: $0.909243\le s\le1$ \\
    \bottomrule
  \end{tabular}
  \par\medskip
  \textbf{(b) Gold-label balance}\par\smallskip
  \begin{tabular}{llcccc}
    \toprule
    \textbf{Source} & \textbf{Binning} & $K=2$ & $K=5$ & $K=10$ & $K=14$ \\
    \midrule
    \multirow{2}{*}{RealToxicity} & EF & 700--700 / 100.0\% & 277--283 / 100.0\% & 130--148 / 99.9\% & 100--100 / 100.0\% \\
    & EW & 468--932 / 94.6\% & 165--588 / 87.4\% & 81--364 / 86.0\% & 41--256 / 84.0\% \\
    \multirow{2}{*}{Civil Comments} & EF & 632--768 / 99.5\% & 219--317 / 99.2\% & 44--191 / 94.5\% & 32--168 / 92.7\% \\
    & EW & 618--782 / 99.3\% & 185--320 / 98.2\% & 15--303 / 84.7\% & 2--198 / 79.6\% \\
    \multirow{2}{*}{STS Benchmark} & EF & 694--706 / 100.0\% & 260--307 / 99.8\% & 124--183 / 99.1\% & 61--123 / 98.5\% \\
    & EW & 633--767 / 99.5\% & 250--362 / 99.0\% & 93--181 / 98.1\% & 63--173 / 95.8\% \\
    \bottomrule
  \end{tabular}
  \caption{\label{tab:candidate-construction-details}
    \textbf{Ordinal candidate construction.} (a) Civil Comments boundary examples; prompts include all $K$ levels in random order. (b) Minimum--maximum items per level / gold coverage $G=\exp(H(q))/K$, with 1,400 items per condition. EF/EW: equal-frequency/equal-width.
  }
  \par\medskip
  \centering
  \scriptsize
  \setlength{\tabcolsep}{2.5pt}
  \begin{minipage}{0.49\textwidth}
  \centering
  \begin{tabular}{lccccccc}
    \toprule
    $K$ & 2 & 3 & 4 & 5 & 6 & 7 & 8 \\
    \midrule
    Ideal $N/K$ & 700 & 466.7 & 350 & 280 & 233.3 & 200 & 175 \\
    Observed & 700 & 466--467 & 350 & 280 & 233--234 & 200 & 175 \\
    \bottomrule
  \end{tabular}
  \end{minipage}\hfill
  \begin{minipage}{0.49\textwidth}
  \centering
  \begin{tabular}{lcccccc}
    \toprule
    $K$ & 9 & 10 & 11 & 12 & 13 & 14 \\
    \midrule
    Ideal $N/K$ & 155.6 & 140 & 127.3 & 116.7 & 107.7 & 100 \\
    Observed & 155--156 & 140 & 127--128 & 116--117 & 107--108 & 100 \\
    \bottomrule
  \end{tabular}
  \end{minipage}
  \caption{\label{tab:gold-position-balance}
    \textbf{Gold-position counts} across three ordinal sources and both binnings; 1,400 items per condition. Ranges show minimum--maximum counts per position.
  }
\end{table*}

\paragraph{Binning rules.}
For a continuous source score $s_i\in[s_{\min},s_{\max}]$, the internal thresholds for equal-width (EW) and equal-frequency (EF) binning are
\begin{equation}
\label{eq:binning}
\begin{array}{@{}l@{}}
t_j^{\mathrm{EW}} = \displaystyle s_{\min}+\frac{j}{K}(s_{\max}-s_{\min}),\\[5pt]
t_j^{\mathrm{EF}} = \displaystyle Q\!\left(\frac{j}{K}\right),\\[5pt]
y_i^{(m)} = \displaystyle 1+\sum_{j=1}^{K-1}{\bf 1}[s_i\ge t_j^{(m)}].
\end{array}
\end{equation}
Here $m\in\{\mathrm{EW},\mathrm{EF}\}$, $j=1,\ldots,K-1$, and $Q$ is the empirical quantile function of the frozen 5,000-item source sample. Equal-frequency thresholds are adjusted when necessary so that identical source scores are not split across levels.

\paragraph{Tie handling and interval closure.}
Equal-width thresholds require no data-dependent adjustment. For equal-frequency binning, each target cut $jN/K$ is placed between adjacent distinct observed scores so that the resulting cumulative count is as close as possible to the target; the recorded threshold is the first score assigned to the upper interval. Consequently, repeated scores always remain in the same level. Every interval is left-closed and right-open, $[t_{j-1},t_j)$, except the final interval, $[t_{K-1},t_K]$, which includes the maximum. A score exactly equal to an internal threshold therefore enters the higher level, as reflected by the $\ge$ operator in Equation~\ref{eq:binning}. We verified strictly increasing edges and non-empty levels for all 78 ordinal conditions (three sources, two binning rules, and 13 values of $K$); no empty-bin repair or post-hoc level merging was applied.

\paragraph{Candidate strings and gold-label balance.}
Table~\ref{tab:candidate-construction-details}(a) illustrates how the same continuous scale is converted into increasingly fine-grained candidate sets. Equal-width binning produces uniformly spaced intervals, whereas equal-frequency binning adapts its boundaries to the empirical score distribution. In both cases, all $K$ candidates are retained in the prompt and only their presentation order is randomized. Table~\ref{tab:candidate-construction-details}(b) reports the resulting gold-label distributions; each cell gives the smallest and largest per-level counts followed by $G=\exp(H(q))/K$. Equal-frequency binning remains close to balanced, with gold effective-label coverage $G$ at or above 92.7\% in every representative condition. Equal-width binning preserves the natural source distribution and is consequently more imbalanced, reaching $G=79.6\%$ for Civil Comments at $K=14$. The released JSONL records contain the complete counts and bin edges for every $K=2,\ldots,14$.

\paragraph{Gold-position balance.}
Candidate order is randomized separately for every item after canonical levels are constructed. Within each source, binning rule, and $K$ condition, the 1,400 evaluated items are allocated across the $K$ presentation positions as evenly as integer counts permit. Table~\ref{tab:gold-position-balance} confirms that no condition differs by more than one item between positions. The gold-label differences in Table~\ref{tab:candidate-construction-details}(b) therefore reflect the binning rule and the source-score distribution and cannot be attributed to gold position, which is balanced by construction.

\paragraph{DBpedia reference.}
Each DBpedia item receives its own random class subset (e.g., 91 distinct subsets at $K=2$ and 952 at $K=5$), so most subsets contain only a handful of items.
We therefore compute $D_R$ within subsets and weight subsets by their size, and we do not report subset-level $U_{\mathrm{arg}}$.
At every $K$, each of the nine models makes at most 15 errors on the 1,400 DBpedia items.

\subsection{Fixed-Count Ordinal Spacing}
\label{sec:appendix-distance-construction}

\paragraph{Source items and exact balance.}
We begin with the frozen 5,000-item samples of Word Concreteness, WMT20 Translation Quality, Measuring Hate Speech, and Stanford Politeness.
All rows sharing an identical context within a source are removed; this excludes 19 Politeness rows and none from the other sources.
Within each original gold level, we retain the same number of items, rounded down to a multiple of $K!$ to balance every candidate ordering.
This removes a further 32, 32, 8, and 67 rows, respectively.
Selection sorts source identifiers and then shuffles each stratum with a SHA-256-derived random generator keyed by seed 20261007, source, and gold level.
Table~\ref{tab:distance-construction} records the resulting sample sizes and spacing values.
These four sources are distinct from the scale-refinement sources: they test ordinal spacing at fixed $K$, rather than introducing new bins on the same continuum.

\paragraph{Distance definition and partition selection.}
Let the original scale have $K_0$ levels, indexed from zero, and let $S$ contain $K$ displayed levels.
Its normalized mean pairwise rank distance is
\begin{equation}
\label{eq:appendix-rank-distance}
d(S)=\frac{2\sum_{a,b\in S,\,a<b}|a-b|}
{K(K-1)(K_0-1)}.
\end{equation}
We enumerate all partitions of the original scale into disjoint sets of size $K$.
For each attainable mean of $d(S)$ over the partition, we retain the partition with the smallest standard deviation of its group distances; remaining ties use lexicographic order.
We then choose seven distinct attainable means nearest to seven equally spaced targets between the minimum and maximum, breaking equal-distance ties toward the smaller mean.
Sorting the selected means gives conditions D1--D7.
Every item receives the group containing its original gold level.
Consecutive groups define D1, and widely interleaved groups define D7.
The seven condition means are increasing, but an individual item's gap need not increase at every step because its group membership changes.

\begin{table*}[t]
  \centering
  \small
  \setlength{\tabcolsep}{4pt}
  \begin{tabular}{lrrrrrrrrrr}
    \toprule
    & & & & \multicolumn{7}{c}{\textbf{Mean normalized ordinal-rank distance}} \\
    \cmidrule(lr){5-11}
    \textbf{Dataset} & $K_0$ & $K$ & $N$/condition & D1 & D2 & D3 & D4 & D5 & D6 & D7 \\
    \midrule
    Word Concreteness & 9 & 3 & 4,968 & .167 & .222 & .278 & .333 & .389 & .444 & .500 \\
    WMT20 Translation Quality & 12 & 3 & 4,968 & .121 & .182 & .242 & .303 & .364 & .424 & .485 \\
    Measuring Hate Speech & 8 & 4 & 4,992 & .238 & .310 & .357 & .381 & .405 & .429 & .476 \\
    Stanford Politeness & 9 & 3 & 4,914 & .167 & .222 & .278 & .333 & .389 & .444 & .500 \\
    \bottomrule
  \end{tabular}
  \caption{\label{tab:distance-construction}
    \textbf{Fixed-count spacing.} $K_0$: original levels; $K$: displayed candidates; $N$: fixed items. D1--D7 report mean normalized rank distances (Equation~\ref{eq:appendix-rank-distance}).
  }
\end{table*}

\paragraph{Fixed variables and ordering.}
The context, question, original correct label, displayed gold position, and $K$ are identical across an item's seven versions.
Within each original-gold stratum, we cycle through all $K!$ orderings, represented by a gold slot and a permutation of the distractors.
The same ordering code is reused across D1--D7.
Consequently, original gold levels, displayed gold positions, and relative gold ranks are balanced in each condition, and every candidate in each ordered set occurs equally often as gold.
An item's relative gold rank can change across conditions; its displayed position cannot.
For example, an item with gold \texttt{Level 05} on a nine-level scale can receive \{\texttt{Level 04}, \texttt{Level 05}, \texttt{Level 06}\} at D1 and \{\texttt{Level 02}, \texttt{Level 05}, \texttt{Level 08}\} at D7, with the correct candidate kept in the same display slot.
No LLM generation or manual rewriting is used in this construction.

\paragraph{Scope of the manipulation.}
Spacing is measured in original ordinal ranks, not raw-score units or embedding distance.
The experiment retains the original target property and level wording while changing the available distractor levels.
It therefore tests whether larger gaps between ordinal alternatives restore utilization, without treating rank distance as a validated general measure of semantic coupling.
All KEV and StartLux checkpoints cover 138,894 queries; JEV covers the 104,118 queries from WMT20, Hate Speech, and Politeness.
Appendix~\ref{sec:appendix-distance-results} reports these panels separately.

\section{Bias-Aware Post-Training Configuration}
\label{sec:appendix-posttraining}

\paragraph{Training-source selection.}
Before adaptation, we fixed eight sources from the main panel using two requirements: each had a public labeled training split, and KEV-0.8B had baseline $D_R>30$ points on the frozen evaluation sample. For all eight sources, predicted effective support was below gold effective support, so these deviations represent compression.
The selected set spans ordinal scales from $K=5$ to $K=14$ and several decision domains: Civil Comments Toxicity, Word Concreteness, WMT20 Translation Quality, Measuring Hate Speech, RealToxicity Continuation, Wine Quality, IBM Argument Quality, and HelpSteer2 Correctness.
The frozen training set contains 4,000 source-indexed examples per dataset (32,000 total), all used for optimization; no validation split was constructed.
The source list and record identities were fixed before any post-training run, and post-training outcomes were not used to select sources.
Baseline performance on the 40-dataset panel determines source eligibility; its records are not used for gradient updates, hyperparameter tuning, or checkpoint selection.
The held-out evaluation items are disjoint from the post-training set by both source ID and normalized-text hash.

\begin{table}[t]
  \centering
  \small
  \setlength{\tabcolsep}{4pt}
  \begin{tabular}{lcc}
    \toprule
    \textbf{Configuration} & \textbf{KEV-0.8B} & \textbf{KEV-4B} \\
    \midrule
    Initialization & KEV-0.8B & KEV-4B \\
    LoRA rank / $\alpha$ / dropout & 16 / 32 / 0.05 & 16 / 32 / 0.05 \\
    Learning rate & $2\times10^{-5}$ & $1.5\times10^{-5}$ \\
    Micro-batch size & 32 & 8 \\
    Gradient accumulation & 2 & 8 \\
    Effective batch size & 64 & 64 \\
    Epochs / optimizer steps & 2 / 1,000 & 2 / 1,000 \\
    State / packed-token cap & 2,048 / 3,712 & 2,048 / 3,712 \\
    Random seed & 42 & 42 \\
    \bottomrule
  \end{tabular}
  \caption{\label{tab:posttraining-config}
    \textbf{BA-LoRA configuration by model size.}
  }
\end{table}

\paragraph{Adapter placement and precision.}
We warm-start the released KEV adapters and pointer heads while keeping the Qwen3.5 backbone frozen.
Rank-16 LoRA adapters \citep{hu2022lora} are applied to all supported attention, MLP, and hybrid projection layers: \texttt{q\_proj}, \texttt{k\_proj}, \texttt{v\_proj}, \texttt{o\_proj}, \texttt{gate\_proj}, \texttt{up\_proj}, \texttt{down\_proj}, \texttt{in\_proj\_qkv}, \texttt{in\_proj\_z}, \texttt{in\_proj\_a}, \texttt{in\_proj\_b}, and \texttt{out\_proj}.
Forward passes and frozen backbone weights use bfloat16, while the LoRA parameters and pointer head retain FP32 master weights.
Gradient checkpointing is disabled for these runs.
The 2,048-token state cap implies a 2,688-token branch cap and a 3,712-token packed-request cap in the KEV training context.

\paragraph{Optimization and BA-LoRA objective.}
We use AdamW with weight decay 0.01, global gradient clipping at 1.0, and a OneCycle learning-rate schedule with 10\% warm-up.
The bias-aware objective augments the supervised decision loss with frozen-original-KEV consistency ($\lambda_{\mathrm{cons}}=0.025$, temperature 2.0), diversity ($\lambda_{\mathrm{div}}=0.005$), and SVD ($\lambda_{\mathrm{svd}}=0.005$) terms.
Consistency follows a cosine decay; diversity and SVD use a two-phase schedule with a 20\% warm-up and 5\% ramp-up.
The SVD term uses the leading 10 singular values, is normalized by the Frobenius norm, and is computed within groups sharing the same candidate count $K$.
We do not add a separate permutation-KL loss.

\paragraph{Model-specific settings and checkpoint selection.}
We did not tune either model on the frozen 40-dataset evaluation panel and did not conduct a model-specific hyperparameter search.
The BA-LoRA loss weights, rank, target modules, seed, precision, effective batch size, and two-epoch duration are shared across sizes; the learning rate and micro-batch/accumulation pair are set separately to accommodate model scale while preserving an effective batch of 64.
Because no validation set or early stopping is used, the reported checkpoint is the deterministic final checkpoint after two epochs (1,000 optimizer updates), rather than the best checkpoint selected from evaluation performance.

\paragraph{Recorded training resources and environment.}
The two BA-LoRA runs use one CUDA process on one GPU each and take 1.56 hours (0.8B) and 5.89 hours (4B), or 7.46 single-GPU hours combined. The training host metadata identifies NVIDIA GeForce RTX 4090 GPUs (48~GB reported per device). The runs recorded Python~3.13.11, CUDA~12.4, PyTorch~2.8.0, Transformers~5.17.0, and PEFT~0.21.0; the released analysis environment specifies Python~3.13 and pinned package versions. BA-LoRA uses one training seed (42) per model size.

\section{Additional Results}
\label{sec:appendix-results}

\subsection{Full-Panel Utilization and Anchor Direction}
\label{sec:appendix-main-results}

\begin{figure*}[t]
  \centering
  \includegraphics[width=\textwidth]{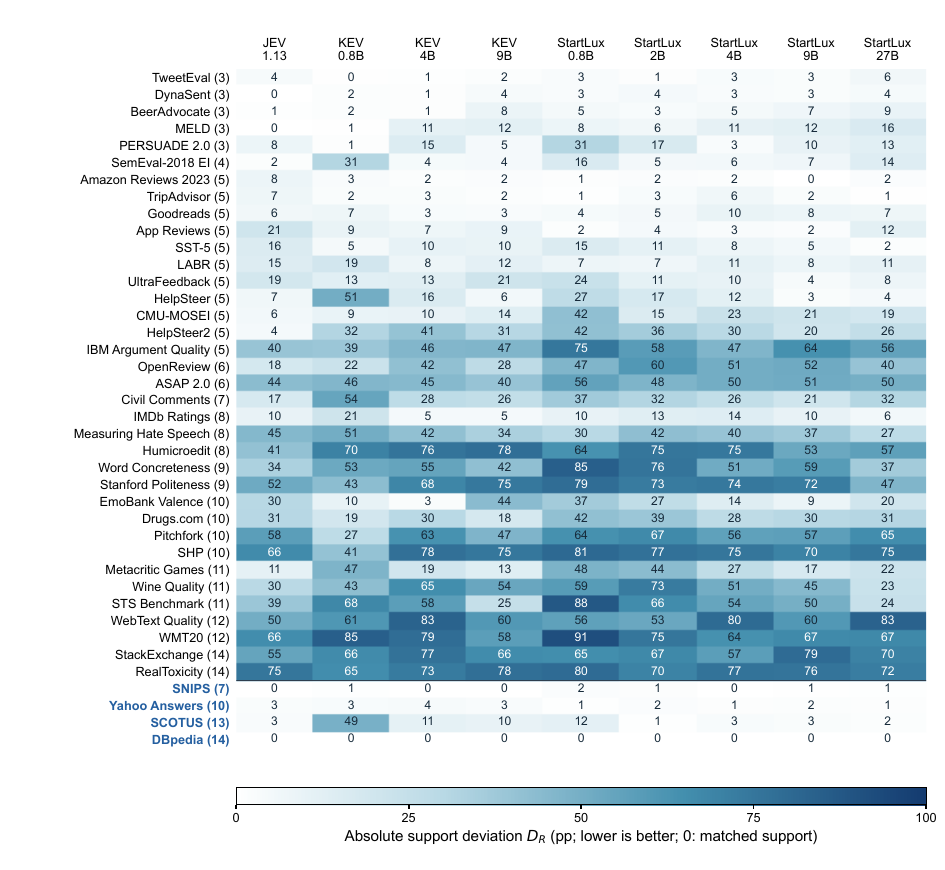}
  \caption{\label{fig:main-detail}
    \textbf{Per-dataset support deviation.} Cells: $D_R$ in pp, rounded to integers. Rows are ordered by $K$ (after each name), then mean $D_R$; the line separates ordinal and nominal tasks.
  }
\end{figure*}

\begin{figure}[t]
  \centering
  \includegraphics[width=\columnwidth]{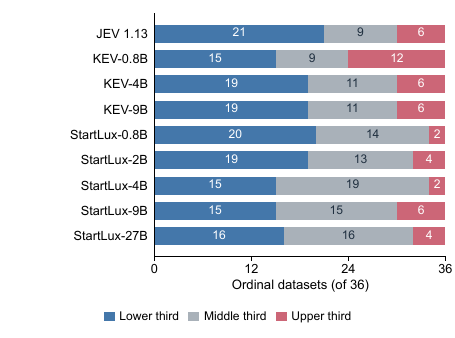}
  \caption{\label{fig:anchor-direction}
    \textbf{Modal prediction location.} Counts of the 36 ordinal datasets whose modal prediction falls in the lower, middle, or upper third (thresholds: Appendix~\ref{sec:appendix-results}).
  }
\end{figure}

Figure~\ref{fig:main-detail} reports $D_R$ for all 40 datasets and all nine models in the main evaluation (\S\ref{sec:broad}).
Figure~\ref{fig:anchor-direction} summarizes where each model's modal prediction lies on the 36 ordinal scales.
Higher values indicate greater effective-support mismatch; the accompanying modal-rank summary describes the direction of label preference.

\paragraph{Compression direction.}
The signed effective-support comparison shows compression on 35 of 36 ordinal datasets for JEV~1.13 and KEV-9B, and on all 36 for KEV-0.8B, KEV-4B, and the five StartLux models. Figure~\ref{fig:main-detail} instead displays the absolute deviation $D_R$.

\paragraph{Cross-dataset variation.}
Across the 36 ordinal datasets, $D_R$ exceeds 30 points on 14--21 datasets per model.
The association between candidate count and compression spans sentiment, toxicity, ratings, argument and response quality, semantic similarity, politeness, and translation quality.
The four nominal tasks have mean TVD of 4.7--18.6\%, compared with 29.8--43.0\% for ordinal tasks (Table~\ref{tab:main}).
Nominal tasks are not uniformly free of concentration.
On SCOTUS, KEV-0.8B assigns 55\% of predictions to the broad ``Miscellaneous'' class, yielding $D_R\approx49$ points.

\paragraph{Anchor location.}
For each ordinal dataset, we take the most frequently predicted level.
With levels indexed $i=0,\dots,K-1$ from lowest to highest, we call it lower if $i<K/3$, upper if $i\ge 2K/3$, and middle otherwise.
The mode is lower/middle/upper on 21/9/6, 15/9/12, 19/11/6, and 19/11/6 datasets for JEV~1.13, KEV-0.8B, KEV-4B, and KEV-9B, respectively. The corresponding counts for StartLux-0.8B, 2B, 4B, 9B, and 27B are 20/14/2, 19/13/4, 15/19/2, 15/15/6, and 16/16/4.

\paragraph{Model-size trends.}
Table~\ref{tab:main} reports accuracy, nMAE, QWK, and utilization for every checkpoint; Section~\ref{sec:broad} compares the KEV and StartLux trends.
Within StartLux, $D_R$ falls from 39.5 points at 0.8B to 29.3 points at 27B as accuracy rises from 31.7\% to 37.2\%.

\subsection{Candidate-Order Diagnostics}
\label{sec:appendix-order-results}

\paragraph{Probability shifts and residual compression.}
In Table~\ref{tab:position}, KEV-4B and StartLux-0.8B mark the lower and upper endpoints of randomized ordinal $D_R$, at 26.1 and 46.0 points, respectively.
Randomization reduces aggregate concentration, but the remaining deviation is much larger than on nominal tasks.

\paragraph{Item-level sensitivity and aggregate utilization.}
Order sensitivity concerns changes to individual answers; $D_R$ concerns effective support across items.
Appendix~\ref{sec:appendix-metrics} defines these distinct quantities and the separate-permutation aggregation used in Table~\ref{tab:position}.
Each randomized order contributes its own dataset-level $D_R$ before averaging; predictions from the two orders are not pooled into a single distribution.

\subsection{Scale-Refinement Endpoints and Diagnostics}
\label{sec:appendix-k-results}

Table~\ref{tab:kcurve-endpoints} gives the equal-frequency $K=2$ and $K=14$ endpoints for all nine models under the same item and binning protocol.

\begin{table*}[t]
  \centering
  \small
  \setlength{\tabcolsep}{7pt}
  \begin{tabular}{lcccc}
    \toprule
    \textbf{Model} & \textbf{Acc: $K=2\rightarrow14$} & \textbf{$D_R$: $K=2\rightarrow14$} & \textbf{TVD: $K=2\rightarrow14$} & \textbf{$\Delta D_R$} \\
    \midrule
    JEV~1.13       & 70.88 $\rightarrow$ 13.55 & 4.14 $\rightarrow$ 45.55 & 14.55 $\rightarrow$ 45.62 & $+41.41$ \\
    KEV-0.8B       & 55.12 $\rightarrow$ 8.74  & 22.66 $\rightarrow$ 73.53 & 33.98 $\rightarrow$ 65.81 & $+50.87$ \\
    KEV-4B         & 70.43 $\rightarrow$ 12.36 & 2.15 $\rightarrow$ 31.57 & 12.10 $\rightarrow$ 33.79 & $+29.42$ \\
    KEV-9B         & 69.29 $\rightarrow$ 12.17 & 3.91 $\rightarrow$ 25.08 & 14.10 $\rightarrow$ 31.38 & $+21.16$ \\
    \midrule
    StartLux-0.8B  & 75.76 $\rightarrow$ 14.83 & 2.57 $\rightarrow$ 23.42 & 11.33 $\rightarrow$ 30.26 & $+20.85$ \\
    StartLux-2B    & 71.38 $\rightarrow$ 14.19 & 7.27 $\rightarrow$ 21.59 & 16.67 $\rightarrow$ 29.26 & $+14.32$ \\
    StartLux-4B    & 76.83 $\rightarrow$ 15.55 & 4.14 $\rightarrow$ 43.90 & 12.36 $\rightarrow$ 41.64 & $+39.75$ \\
    StartLux-9B    & 78.62 $\rightarrow$ 17.81 & 2.19 $\rightarrow$ 37.75 & 6.81 $\rightarrow$ 35.86  & $+35.56$ \\
    StartLux-27B   & 78.76 $\rightarrow$ 16.48 & 4.03 $\rightarrow$ 46.47 & 11.71 $\rightarrow$ 43.71 & $+42.45$ \\
    \bottomrule
  \end{tabular}
  \caption{\label{tab:kcurve-endpoints}
    \textbf{Equal-frequency endpoints.} Macro-averages over three datasets (1,400 fixed items each). Acc/TVD: \%; $D_R$ and $\Delta D_R=D_R(14)-D_R(2)$: pp. Curves: Figure~\ref{fig:kcurve}.
  }
\end{table*}

\paragraph{Binning-specific variation.}
All nine models have higher endpoint $D_R$ at $K=14$ than at $K=2$ under both binning rules (Figure~\ref{fig:kcurve}, upper row).
Equal-width curves are less regular than equal-frequency curves, and the effect size depends on the construction.
Section~\ref{sec:kscaling} illustrates this construction dependence with the KEV-9B endpoints.
Thus, endpoint increases need not imply monotonic increases at every intermediate $K$.

\paragraph{Probability and ordinal-error diagnostics.}
Under equal-frequency binning, KEV-0.8B's mean top-1 margin falls from 33.8 to 13.7 points, while $U_{\mathrm{soft}}$ falls from 95.5\% to 85.6\%.
JEV~1.13's margin falls from 69.0 to 21.8 points, alongside decreasing soft utilization (Figure~\ref{fig:kcurve}, lower row).
These models differ from KEV-4B and KEV-9B, whose soft utilization remains at least 96\% under both binnings.
For JEV and KEV, endpoint nMAE remains comparatively stable or improves as $K$ increases, despite the growth in $D_R$.
For example, JEV~1.13's equal-width QWK rises from 0.437 to 0.504.

\paragraph{DBpedia ceiling reference.}
JEV~1.13's top-1 margin remains above 96 points throughout the DBpedia candidate-count sweep.
This complements its high accuracy and low support deviation reported in Section~\ref{sec:kscaling}.
DBpedia is a high-accuracy ceiling reference, not a difficulty-matched control; Appendix~\ref{sec:appendix-k-construction} details its candidate subsets and aggregation.

\subsection{Fixed-Count Spacing Results}
\label{sec:appendix-distance-results}

Tables~\ref{tab:distance-wmt}, \ref{tab:distance-hate}, and~\ref{tab:distance-politeness} report all seven conditions for the three datasets shared by all nine models (\S\ref{sec:distance}).
Table~\ref{tab:distance-word} gives the corresponding Word Concreteness results for the eight open checkpoints.
Each curve reuses exactly the same items and fixed gold display positions throughout D1--D7.
Metrics use relative ordinal rank within the displayed candidate set, as defined in Appendix~\ref{sec:appendix-metrics}; they are not distributions over A/B/C/D positions or over the original full-scale labels.
The complete condition sequence preserves intermediate reversals that endpoint comparisons alone can conceal.
The condition-level sample size and gold-rank distribution are constant throughout each dataset's curve.
The dataset-level tables are primary: a three-source macro-average can combine different utilization trajectories.

\begin{table*}[t]
  \centering
  \small
  \setlength{\tabcolsep}{5pt}
  \begin{tabular}{llrrrrrrr}
    \toprule
    \textbf{Model} & \textbf{Metric} & D1 & D2 & D3 & D4 & D5 & D6 & D7 \\
    \midrule
    JEV~1.13 & Acc $\uparrow$ & 34.34 & 34.00 & 35.45 & 35.47 & 36.23 & 36.37 & 36.41 \\
     & $D_R$ $\downarrow$ & 23.78 & 21.16 & 20.65 & 17.90 & 16.29 & 14.43 & 13.24 \\
     & TVD $\downarrow$ & 33.15 & 31.80 & 31.54 & 30.35 & 29.13 & 26.81 & 24.80 \\
    \cmidrule(lr){1-9}
    KEV-0.8B & Acc $\uparrow$ & 33.84 & 33.92 & 34.16 & 33.88 & 34.34 & 34.34 & 33.55 \\
     & $D_R$ $\downarrow$ & 16.86 & 5.91 & 6.31 & 6.52 & 0.75 & 1.62 & 1.36 \\
     & TVD $\downarrow$ & 27.84 & 16.91 & 15.98 & 16.73 & 5.33 & 8.25 & 7.27 \\
    \cmidrule(lr){1-9}
    KEV-4B & Acc $\uparrow$ & 33.37 & 32.91 & 33.62 & 34.28 & 34.84 & 34.34 & 34.30 \\
     & $D_R$ $\downarrow$ & 5.20 & 3.76 & 5.87 & 6.98 & 11.01 & 10.53 & 8.64 \\
     & TVD $\downarrow$ & 14.39 & 11.74 & 14.51 & 15.42 & 23.25 & 20.51 & 18.64 \\
    \cmidrule(lr){1-9}
    KEV-9B & Acc $\uparrow$ & 33.49 & 34.08 & 35.25 & 34.60 & 35.02 & 34.94 & 35.04 \\
     & $D_R$ $\downarrow$ & 13.68 & 14.49 & 22.15 & 21.93 & 12.75 & 19.26 & 24.39 \\
     & TVD $\downarrow$ & 26.55 & 27.38 & 31.78 & 33.82 & 21.74 & 27.03 & 29.27 \\
    \cmidrule(lr){1-9}
    StartLux-0.8B & Acc $\uparrow$ & 33.15 & 33.66 & 32.67 & 33.13 & 32.53 & 33.39 & 33.39 \\
     & $D_R$ $\downarrow$ & 60.30 & 23.29 & 37.41 & 40.42 & 41.84 & 62.46 & 65.42 \\
     & TVD $\downarrow$ & 63.12 & 35.47 & 45.97 & 46.78 & 43.56 & 64.51 & 66.12 \\
    \cmidrule(lr){1-9}
    StartLux-2B & Acc $\uparrow$ & 33.21 & 33.31 & 33.88 & 34.20 & 34.04 & 34.16 & 34.02 \\
     & $D_R$ $\downarrow$ & 7.14 & 9.34 & 9.39 & 15.35 & 14.13 & 11.48 & 13.64 \\
     & TVD $\downarrow$ & 17.37 & 18.06 & 17.69 & 28.20 & 26.57 & 19.77 & 22.12 \\
    \cmidrule(lr){1-9}
    StartLux-4B & Acc $\uparrow$ & 31.88 & 32.45 & 33.19 & 33.84 & 34.50 & 34.92 & 35.97 \\
     & $D_R$ $\downarrow$ & 14.97 & 23.22 & 17.95 & 19.32 & 23.47 & 25.23 & 26.82 \\
     & TVD $\downarrow$ & 27.70 & 35.12 & 30.84 & 32.00 & 35.77 & 37.36 & 38.71 \\
    \cmidrule(lr){1-9}
    StartLux-9B & Acc $\uparrow$ & 31.52 & 31.34 & 32.13 & 32.37 & 32.67 & 33.09 & 33.31 \\
     & $D_R$ $\downarrow$ & 6.47 & 7.33 & 3.65 & 9.40 & 10.20 & 15.55 & 12.08 \\
     & TVD $\downarrow$ & 16.26 & 16.87 & 11.07 & 21.10 & 22.42 & 27.44 & 24.24 \\
    \cmidrule(lr){1-9}
    StartLux-27B & Acc $\uparrow$ & 34.10 & 33.80 & 34.68 & 34.44 & 36.07 & 35.37 & 35.93 \\
     & $D_R$ $\downarrow$ & 20.88 & 18.48 & 21.75 & 18.75 & 15.48 & 14.49 & 12.99 \\
     & TVD $\downarrow$ & 32.00 & 30.35 & 32.79 & 31.30 & 28.40 & 27.01 & 21.50 \\
    \bottomrule
  \end{tabular}
  \caption{\label{tab:distance-wmt}
    \textbf{WMT20 spacing} (4,968 fixed items, $K=3$). Acc/TVD: \%; $D_R$: pp. Distributions use relative ordinal ranks, not display positions. D1--D7: Table~\ref{tab:distance-construction}.
  }
\end{table*}

\begin{table*}[t]
  \centering
  \small
  \setlength{\tabcolsep}{5pt}
  \begin{tabular}{llrrrrrrr}
    \toprule
    \textbf{Model} & \textbf{Metric} & D1 & D2 & D3 & D4 & D5 & D6 & D7 \\
    \midrule
    JEV~1.13 & Acc $\uparrow$ & 36.30 & 40.18 & 41.41 & 43.07 & 43.71 & 42.63 & 48.08 \\
     & $D_R$ $\downarrow$ & 19.95 & 19.05 & 17.40 & 19.84 & 14.44 & 8.29 & 8.76 \\
     & TVD $\downarrow$ & 29.43 & 29.81 & 26.92 & 27.04 & 25.50 & 19.13 & 19.67 \\
    \cmidrule(lr){1-9}
    KEV-0.8B & Acc $\uparrow$ & 27.58 & 29.35 & 29.39 & 32.85 & 29.67 & 30.49 & 31.87 \\
     & $D_R$ $\downarrow$ & 16.85 & 21.65 & 21.03 & 21.15 & 20.05 & 16.00 & 15.01 \\
     & TVD $\downarrow$ & 28.02 & 32.63 & 32.25 & 32.45 & 30.79 & 25.62 & 26.04 \\
    \cmidrule(lr){1-9}
    KEV-4B & Acc $\uparrow$ & 33.83 & 35.42 & 36.98 & 37.70 & 38.84 & 38.42 & 42.89 \\
     & $D_R$ $\downarrow$ & 12.89 & 12.71 & 16.00 & 14.74 & 15.05 & 12.80 & 6.15 \\
     & TVD $\downarrow$ & 22.76 & 23.00 & 25.64 & 25.64 & 23.72 & 23.34 & 14.68 \\
    \cmidrule(lr){1-9}
    KEV-9B & Acc $\uparrow$ & 32.95 & 36.54 & 37.06 & 36.48 & 40.42 & 39.90 & 42.81 \\
     & $D_R$ $\downarrow$ & 17.63 & 14.02 & 10.15 & 11.76 & 8.79 & 9.37 & 13.87 \\
     & TVD $\downarrow$ & 26.94 & 25.28 & 20.93 & 20.77 & 18.55 & 18.33 & 20.85 \\
    \cmidrule(lr){1-9}
    StartLux-0.8B & Acc $\uparrow$ & 36.88 & 38.80 & 43.31 & 44.87 & 45.81 & 45.31 & 48.62 \\
     & $D_R$ $\downarrow$ & 9.78 & 9.70 & 18.12 & 21.00 & 25.26 & 23.74 & 9.39 \\
     & TVD $\downarrow$ & 17.15 & 17.01 & 24.80 & 24.34 & 32.15 & 34.70 & 16.39 \\
    \cmidrule(lr){1-9}
    StartLux-2B & Acc $\uparrow$ & 38.30 & 42.05 & 44.25 & 46.25 & 45.31 & 45.01 & 49.88 \\
     & $D_R$ $\downarrow$ & 20.64 & 14.50 & 13.36 & 15.97 & 11.21 & 13.17 & 7.39 \\
     & TVD $\downarrow$ & 27.38 & 21.59 & 19.35 & 25.26 & 18.19 & 19.31 & 17.11 \\
    \cmidrule(lr){1-9}
    StartLux-4B & Acc $\uparrow$ & 34.94 & 38.10 & 39.98 & 41.79 & 41.25 & 41.59 & 45.53 \\
     & $D_R$ $\downarrow$ & 22.10 & 21.00 & 20.60 & 20.22 & 19.49 & 15.46 & 13.08 \\
     & TVD $\downarrow$ & 32.55 & 31.41 & 30.09 & 30.97 & 28.45 & 25.38 & 24.34 \\
    \cmidrule(lr){1-9}
    StartLux-9B & Acc $\uparrow$ & 37.16 & 40.85 & 42.91 & 42.99 & 45.25 & 44.27 & 50.06 \\
     & $D_R$ $\downarrow$ & 21.65 & 18.89 & 16.11 & 20.50 & 16.10 & 16.01 & 8.99 \\
     & TVD $\downarrow$ & 28.93 & 26.96 & 25.10 & 29.93 & 24.90 & 25.46 & 16.07 \\
    \cmidrule(lr){1-9}
    StartLux-27B & Acc $\uparrow$ & 35.56 & 38.98 & 40.24 & 41.29 & 41.75 & 41.29 & 45.43 \\
     & $D_R$ $\downarrow$ & 21.26 & 20.31 & 18.89 & 20.68 & 14.85 & 13.01 & 12.64 \\
     & TVD $\downarrow$ & 31.81 & 31.37 & 30.19 & 31.17 & 25.18 & 22.52 & 23.04 \\
    \bottomrule
  \end{tabular}
  \caption{\label{tab:distance-hate}
    \textbf{Measuring Hate Speech spacing} (4,992 fixed items, $K=4$). Acc/TVD: \%; $D_R$: pp. Distributions use relative ordinal ranks, not display positions. D1--D7: Table~\ref{tab:distance-construction}.
  }
\end{table*}

\begin{table*}[t]
  \centering
  \small
  \setlength{\tabcolsep}{5pt}
  \begin{tabular}{llrrrrrrr}
    \toprule
    \textbf{Model} & \textbf{Metric} & D1 & D2 & D3 & D4 & D5 & D6 & D7 \\
    \midrule
    JEV~1.13 & Acc $\uparrow$ & 38.22 & 38.22 & 40.58 & 40.39 & 41.80 & 42.04 & 43.61 \\
     & $D_R$ $\downarrow$ & 6.11 & 4.11 & 3.11 & 3.46 & 4.11 & 11.99 & 26.76 \\
     & TVD $\downarrow$ & 16.89 & 13.94 & 11.68 & 11.33 & 13.09 & 21.98 & 34.94 \\
    \cmidrule(lr){1-9}
    KEV-0.8B & Acc $\uparrow$ & 34.98 & 34.82 & 34.64 & 34.64 & 34.33 & 34.45 & 34.80 \\
     & $D_R$ $\downarrow$ & 2.95 & 2.60 & 2.23 & 0.70 & 0.62 & 0.15 & 0.10 \\
     & TVD $\downarrow$ & 10.36 & 9.61 & 9.10 & 5.31 & 4.56 & 2.54 & 1.93 \\
    \cmidrule(lr){1-9}
    KEV-4B & Acc $\uparrow$ & 34.15 & 33.90 & 33.11 & 33.80 & 33.54 & 33.25 & 32.60 \\
     & $D_R$ $\downarrow$ & 13.16 & 16.74 & 21.11 & 22.97 & 27.91 & 29.25 & 26.97 \\
     & TVD $\downarrow$ & 23.71 & 29.02 & 33.74 & 35.39 & 37.55 & 36.91 & 29.57 \\
    \cmidrule(lr){1-9}
    KEV-9B & Acc $\uparrow$ & 33.90 & 34.21 & 34.07 & 34.86 & 35.12 & 34.74 & 34.62 \\
     & $D_R$ $\downarrow$ & 4.57 & 7.55 & 9.54 & 13.49 & 13.29 & 14.21 & 29.97 \\
     & TVD $\downarrow$ & 14.77 & 19.29 & 21.86 & 26.27 & 25.95 & 23.42 & 31.26 \\
    \cmidrule(lr){1-9}
    StartLux-0.8B & Acc $\uparrow$ & 32.44 & 32.74 & 32.93 & 33.19 & 33.56 & 33.29 & 32.40 \\
     & $D_R$ $\downarrow$ & 46.32 & 47.82 & 56.33 & 58.07 & 61.56 & 55.69 & 45.99 \\
     & TVD $\downarrow$ & 53.52 & 54.56 & 60.50 & 61.74 & 63.86 & 58.75 & 48.03 \\
    \cmidrule(lr){1-9}
    StartLux-2B & Acc $\uparrow$ & 32.99 & 32.64 & 33.31 & 33.25 & 32.28 & 33.07 & 32.44 \\
     & $D_R$ $\downarrow$ & 11.96 & 9.94 & 26.81 & 27.98 & 29.05 & 36.70 & 35.76 \\
     & TVD $\downarrow$ & 24.34 & 20.82 & 34.90 & 36.20 & 36.39 & 34.25 & 33.17 \\
    \cmidrule(lr){1-9}
    StartLux-4B & Acc $\uparrow$ & 34.00 & 33.76 & 33.33 & 33.54 & 34.29 & 34.68 & 34.72 \\
     & $D_R$ $\downarrow$ & 26.53 & 22.42 & 25.12 & 32.20 & 33.88 & 35.78 & 33.68 \\
     & TVD $\downarrow$ & 38.46 & 34.74 & 37.02 & 43.06 & 44.14 & 37.93 & 33.03 \\
    \cmidrule(lr){1-9}
    StartLux-9B & Acc $\uparrow$ & 33.19 & 32.97 & 33.01 & 33.17 & 33.01 & 33.82 & 33.86 \\
     & $D_R$ $\downarrow$ & 33.26 & 31.58 & 32.82 & 33.52 & 32.33 & 36.61 & 31.25 \\
     & TVD $\downarrow$ & 41.47 & 40.23 & 40.84 & 41.39 & 42.63 & 41.92 & 32.66 \\
    \cmidrule(lr){1-9}
    StartLux-27B & Acc $\uparrow$ & 32.52 & 32.48 & 32.68 & 32.68 & 33.03 & 33.54 & 33.46 \\
     & $D_R$ $\downarrow$ & 25.56 & 25.24 & 29.56 & 34.88 & 33.97 & 34.09 & 38.11 \\
     & TVD $\downarrow$ & 37.00 & 36.59 & 37.61 & 43.65 & 42.61 & 32.76 & 36.39 \\
    \bottomrule
  \end{tabular}
  \caption{\label{tab:distance-politeness}
    \textbf{Stanford Politeness spacing} (4,914 fixed items, $K=3$). Acc/TVD: \%; $D_R$: pp. Distributions use relative ordinal ranks, not display positions. D1--D7: Table~\ref{tab:distance-construction}.
  }
\end{table*}

\begin{table*}[t]
  \centering
  \small
  \setlength{\tabcolsep}{5pt}
  \begin{tabular}{llrrrrrrr}
    \toprule
    \textbf{Model} & \textbf{Metric} & D1 & D2 & D3 & D4 & D5 & D6 & D7 \\
    \midrule
    KEV-0.8B & Acc $\uparrow$ & 35.14 & 34.30 & 35.37 & 34.86 & 34.74 & 35.61 & 34.30 \\
     & $D_R$ $\downarrow$ & 2.54 & 3.51 & 2.70 & 2.65 & 1.51 & 0.95 & 7.04 \\
     & TVD $\downarrow$ & 10.06 & 12.42 & 11.09 & 10.23 & 7.97 & 5.96 & 18.58 \\
    \cmidrule(lr){1-9}
    KEV-4B & Acc $\uparrow$ & 40.68 & 43.04 & 43.38 & 46.78 & 46.98 & 49.62 & 53.10 \\
     & $D_R$ $\downarrow$ & 15.25 & 11.93 & 13.05 & 14.47 & 18.28 & 16.27 & 15.05 \\
     & TVD $\downarrow$ & 25.60 & 21.90 & 24.03 & 27.33 & 31.02 & 29.09 & 27.74 \\
    \cmidrule(lr){1-9}
    KEV-9B & Acc $\uparrow$ & 41.30 & 42.39 & 45.29 & 49.11 & 49.86 & 53.66 & 56.80 \\
     & $D_R$ $\downarrow$ & 16.37 & 18.45 & 19.50 & 12.83 & 13.61 & 9.04 & 8.72 \\
     & TVD $\downarrow$ & 25.50 & 28.84 & 31.68 & 21.78 & 21.50 & 18.98 & 18.74 \\
    \cmidrule(lr){1-9}
    StartLux-0.8B & Acc $\uparrow$ & 35.45 & 35.31 & 37.22 & 37.48 & 38.04 & 36.07 & 38.02 \\
     & $D_R$ $\downarrow$ & 9.33 & 9.39 & 30.72 & 33.09 & 32.29 & 26.30 & 32.39 \\
     & TVD $\downarrow$ & 19.32 & 21.36 & 41.67 & 43.34 & 41.79 & 30.07 & 32.95 \\
    \cmidrule(lr){1-9}
    StartLux-2B & Acc $\uparrow$ & 40.52 & 38.81 & 40.12 & 39.92 & 39.88 & 44.02 & 39.79 \\
     & $D_R$ $\downarrow$ & 10.07 & 7.77 & 6.93 & 18.65 & 18.88 & 11.19 & 28.75 \\
     & TVD $\downarrow$ & 19.36 & 17.27 & 15.48 & 26.81 & 25.87 & 20.99 & 33.09 \\
    \cmidrule(lr){1-9}
    StartLux-4B & Acc $\uparrow$ & 44.95 & 50.85 & 56.04 & 62.42 & 64.63 & 67.39 & 69.48 \\
     & $D_R$ $\downarrow$ & 0.46 & 0.35 & 0.23 & 2.32 & 2.99 & 2.48 & 1.08 \\
     & TVD $\downarrow$ & 4.47 & 3.60 & 3.14 & 9.84 & 11.33 & 10.39 & 6.34 \\
    \cmidrule(lr){1-9}
    StartLux-9B & Acc $\uparrow$ & 38.75 & 41.47 & 45.87 & 47.48 & 49.03 & 53.86 & 55.35 \\
     & $D_R$ $\downarrow$ & 11.28 & 6.71 & 7.27 & 5.72 & 10.22 & 6.46 & 12.98 \\
     & TVD $\downarrow$ & 23.65 & 18.08 & 15.58 & 14.98 & 20.15 & 15.58 & 21.34 \\
    \cmidrule(lr){1-9}
    StartLux-27B & Acc $\uparrow$ & 43.58 & 47.14 & 51.07 & 53.64 & 56.32 & 61.35 & 65.34 \\
     & $D_R$ $\downarrow$ & 9.19 & 10.03 & 11.45 & 13.51 & 14.83 & 5.06 & 6.69 \\
     & TVD $\downarrow$ & 19.22 & 19.55 & 20.47 & 21.40 & 22.24 & 13.79 & 16.30 \\
    \bottomrule
  \end{tabular}
  \caption{\label{tab:distance-word}
    \textbf{Word Concreteness spacing} (4,968 fixed items, $K=3$). Acc/TVD: \%; $D_R$: pp. Distributions use relative ordinal ranks, not display positions. D1--D7: Table~\ref{tab:distance-construction}. JEV is excluded because its curve is incomplete.
  }
\end{table*}

\paragraph{Non-monotonic and model-specific spacing responses.}
On Politeness, JEV's $D_R$ first falls and then rises, so the endpoint contrast does not describe a monotonic trajectory.
KEV-4B, KEV-9B, and StartLux-2B also have higher $D_R$ and TVD at D7 than at D1.
KEV-0.8B instead approaches matched effective support without moving far from chance-level accuracy, as discussed in Section~\ref{sec:distance} (Table~\ref{tab:distance-politeness}).
These model-specific responses do not follow a consistent size trend.
The intervention changes available distractor levels as well as rank spacing, as specified in Appendix~\ref{sec:appendix-distance-construction}.

\subsection{Full Post-Training Results}
\label{sec:appendix-posttraining-results}

Table~\ref{tab:balora} gives all six metrics before and after BA-LoRA for each evaluation group.
Figure~\ref{fig:ba-lora-radar} in the main text visualizes the eight training-source datasets; the table additionally includes the 28 unseen ordinal sources, all 36 ordinal sources, and the four nominal sources.
Training hyperparameters, adapter modules, source-selection criteria, and checkpoint selection are detailed in Appendix~\ref{sec:appendix-posttraining}.

\begin{table*}[t]
  \centering
  \scriptsize
  \setlength{\tabcolsep}{2.3pt}
  \resizebox{\textwidth}{!}{%
  \begin{tabular}{llrrrrrrrr}
    \toprule
    \textbf{Model} & \textbf{Evaluation group} & $D$ & \textbf{Acc.} $\uparrow$ & $\mathbf{U}_{\mathrm{arg}}$ $\uparrow$ & $\mathbf{U}_{\mathrm{soft}}$ $\uparrow$ & $\mathbf{D_R}$ $\downarrow$ & \textbf{TVD} $\downarrow$ & \textbf{nMAE} $\downarrow$ & \textbf{QWK} $\uparrow$ \\
    \midrule
    KEV-0.8B & Training-source ordinal & 8  & 15.5 $\rightarrow$ 28.5 & 46.3 $\rightarrow$ 85.0 & 96.3 $\rightarrow$ 97.2 & 52.6 $\rightarrow$ 13.2 & 54.7 $\rightarrow$ 19.7 & 37.9 $\rightarrow$ 20.4 & .188 $\rightarrow$ .600 \\
             & Unseen-source ordinal   & 28 & 32.6 $\rightarrow$ 33.4 & 72.8 $\rightarrow$ 69.7 & 97.3 $\rightarrow$ 91.7 & 25.0 $\rightarrow$ 29.9 & 32.5 $\rightarrow$ 35.1 & 28.3 $\rightarrow$ 25.1 & .422 $\rightarrow$ .456 \\
             & All ordinal             & 36 & 28.8 $\rightarrow$ 32.3 & 66.9 $\rightarrow$ 73.1 & 97.1 $\rightarrow$ 92.9 & 31.1 $\rightarrow$ 26.2 & 37.4 $\rightarrow$ 31.7 & 30.5 $\rightarrow$ 24.0 & .370 $\rightarrow$ .488 \\
             & Nominal                 & 4  & 69.6 $\rightarrow$ 72.7 & 83.4 $\rightarrow$ 90.2 & 91.2 $\rightarrow$ 91.1 & 13.4 $\rightarrow$ 4.2 & 18.6 $\rightarrow$ 12.8 & -- & -- \\
    \midrule
    KEV-4B   & Training-source ordinal & 8  & 17.3 $\rightarrow$ 32.7 & 45.7 $\rightarrow$ 83.9 & 90.4 $\rightarrow$ 96.7 & 53.6 $\rightarrow$ 14.4 & 52.9 $\rightarrow$ 19.0 & 32.5 $\rightarrow$ 17.8 & .240 $\rightarrow$ .664 \\
             & Unseen-source ordinal   & 28 & 38.0 $\rightarrow$ 39.4 & 70.9 $\rightarrow$ 78.0 & 94.5 $\rightarrow$ 91.0 & 26.8 $\rightarrow$ 21.3 & 33.1 $\rightarrow$ 27.6 & 22.5 $\rightarrow$ 19.7 & .517 $\rightarrow$ .590 \\
             & All ordinal             & 36 & 33.4 $\rightarrow$ 37.9 & 65.3 $\rightarrow$ 79.3 & 93.6 $\rightarrow$ 92.2 & 32.8 $\rightarrow$ 19.7 & 37.5 $\rightarrow$ 25.7 & 24.7 $\rightarrow$ 19.3 & .455 $\rightarrow$ .606 \\
             & Nominal                 & 4  & 79.7 $\rightarrow$ 80.3 & 94.8 $\rightarrow$ 94.3 & 95.8 $\rightarrow$ 94.5 & 3.8 $\rightarrow$ 4.5 & 9.9 $\rightarrow$ 9.6 & -- & -- \\
    \bottomrule
  \end{tabular}
  }
  \caption{\label{tab:balora}
    \textbf{Before $\rightarrow$ after BA-LoRA.} Macro-averages over $D$ datasets. Units: $D_R$ in pp; QWK unscaled; others in \%. Textual gains are computed before rounding these displayed means. Dashes mark inapplicable ordinal metrics.
  }
\end{table*}

\paragraph{Argmax and soft utilization.}
On the eight training-source datasets, $U_{\mathrm{arg}}$ rises by 38.7 and 38.2 points for KEV-0.8B and KEV-4B, whereas $U_{\mathrm{soft}}$ rises by 0.9 and 6.3 points, respectively (Section~\ref{sec:posttraining}).
Table~\ref{tab:balora} reports the accompanying accuracy, nMAE, and QWK improvements for all 36 ordinal datasets, alongside separate training-source and unseen ordinal-source results.

\paragraph{Nominal transfer.}
Macro-averaged over the four nominal datasets, accuracy and TVD improve at both model sizes (Table~\ref{tab:balora}).
Mean nominal $D_R$ falls from 13.4 to 4.2 points for KEV-0.8B, but increases from 3.8 to 4.5 points for KEV-4B.
The transfer pattern therefore depends on the evaluation group and metric.

\begin{figure*}[t]
  \centering
  \includegraphics[width=\textwidth,height=0.73\textheight,keepaspectratio]{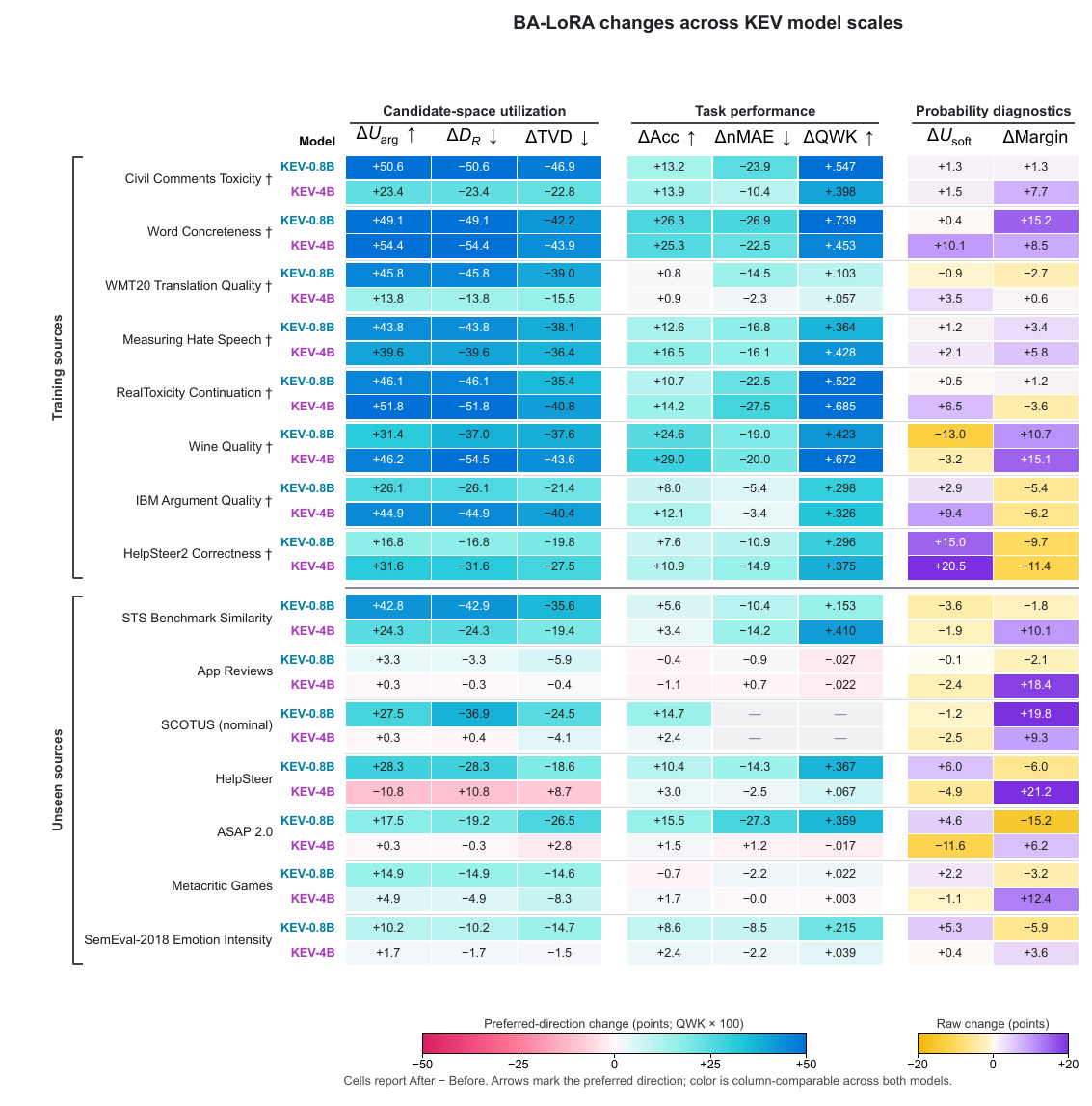}
  \caption{\label{fig:ba_lora_change}
    \textbf{BA-LoRA changes.} Fifteen datasets selected post hoc by KEV-0.8B's largest $D_R$ reductions. Cells: after minus before (QWK unscaled; others in pp). Daggers mark eight training sources; the other seven comprise six ordinal sources and nominal SCOTUS. Group-level macro-averages: Table~\ref{tab:balora}.
  }
\end{figure*}

\paragraph{Selected per-dataset illustration.}
Figure~\ref{fig:ba_lora_change} shows the 15 largest KEV-0.8B reductions in $D_R$ after BA-LoRA.
The eight training-source datasets are Civil Comments Toxicity, Word Concreteness, WMT20 Translation Quality, Measuring Hate Speech, RealToxicity Continuation, Wine Quality, IBM Argument Quality, and HelpSteer2 Correctness.
The seven datasets not used for training comprise six ordinal sources (STS Benchmark Similarity, App Reviews, HelpSteer, ASAP2, Metacritic Games, and SemEval-2018 Emotion Intensity) and one nominal source (SCOTUS).
Because the subset is selected by the observed KEV-0.8B reduction in $D_R$, it is a descriptive view of the strongest improvements rather than an unbiased estimate of transfer.

\end{document}